\documentclass{CUP-JNL-EDS}
\usepackage{graphicx}
\usepackage{algorithm}
\usepackage{algpseudocode}
\usepackage{dirtytalk}
\usepackage{threeparttable}
\usepackage[T1]{fontenc}
\usepackage{times}
\usepackage{newtxmath}
\usepackage[table]{xcolor}  
\usepackage{hyperref}
\usepackage{lipsum}
\usepackage[backend=biber]{biblatex}
\algnewcommand{\LeftComment}[1]{\Statex \(\triangleright\) #1}
\begin{document}

\begin{Frontmatter}
\title[Article Title]{A Temporal Stochastic Bias Correction using a Machine Learning Attention model}

\author[*1]{Omer Nivron}
\author[2]{Damon J. Wischik}
\author[3]{Mathieu Vrac}
\author[4]{Emily Shuckburgh}
\author[4]{Alex T. Archibald}

\address[1]{\orgname{University of Cambridge}, \orgdiv{Department of Computer Science and Technology}, \orgaddress{\city{Cambridge}, \country{UK}}. \texttt{on234@cam.ac.uk}}

\address[2]{ \orgname{University of Cambridge},
\orgdiv{Department of Computer Science and Technology},
\orgaddress{\city{Cambridge}, \country{UK}}}

\address[3]{\orgname{Institut Pierre-Simon Laplace}, \orgdiv{Laboratoire des Sciences du Climat et de l'Environnement}, \orgaddress{\city{Paris}, \country{France}}}

\address[4]{ \orgname{University of Cambridge}, \orgdiv{Department of Computer Science and Technology}, \orgaddress{\city{Cambridge}, \country{UK}}}

\address[4*]{\orgname{University of Cambridge}, \orgdiv{Yusuf Hamied Department of Chemistry},\orgaddress{\city{Cambridge}, \country{UK}}}

\authormark{Nivron et al.}

\keywords{Bias-Correction; Time-series; Heatwaves; Machine-Learning; Attention}

\abstract{Climate models are biased with respect to real-world observations. They usually need to be adjusted before being used in impact studies. The suite of statistical methods that enable such adjustments is called bias correction (BC). However, BC methods currently struggle to adjust temporal biases. Because they mostly disregard the dependence between consecutive time points. As a result, climate statistics with long-range temporal properties, such as heatwave duration and frequency, cannot be corrected accurately. This makes it more difficult to produce reliable impact studies on such climate statistics. This paper offers a novel BC methodology to correct temporal biases. This is made possible by rethinking the philosophy behind BC. We will introduce BC as a time-indexed regression task with stochastic outputs. Rethinking BC enables us to adapt state-of-the-art machine learning (ML) attention models and thereby learn different types of biases, including temporal asynchronicities. With a case study of heatwave duration statistics in Abuja, Nigeria, and Tokyo, Japan, we show more accurate results than current climate model outputs and alternative BC methods.}

\policy{Climate models are biased. Bias Correction (BC) is a suite of statistical methods to address those biases. Corrected climate model statistics enable more precise impact studies. However, to date, BC struggles to adjust for long-range temporal properties such as heatwave duration. We offer a novel methodology for stochastic temporal BC. This is possible by re-thinking BC as a time-indexed regression model with stochastic outputs. This reformulation enables using state-of-the-art machine learning (ML) attention models which can learn different types of biases, including temporal asynchronicities. We show more accurate results than current climate model outputs and alternative BC methods on a  heatwave duration statistics case study in Abuja, Nigeria, and Tokyo, Japan. The results are relevant to a broad audience, ranging from climate scientists to policy-makers.}

\end{Frontmatter}

\section[This is an A Head]{Introduction}
Climate models are biased with respect to observations \cite{PalmerDiag, Christensen2008OnTN, Teutschbein2012BiasCO}, affecting impact studies in domains such as the economy \cite{Tol2018TheEI, /content/publication/9789264235410-en, Tol2021TheEI}, human migration \cite{Dimitri2017}, biodiversity \cite{Bellard2012}, hydrology \cite{Gleick1989ClimateCH, Christensen2004TheEO, Piao2010TheIO} and agronomy \cite{Ciais2005, Ari2020CausesAI}. If left uncorrected, those biases can invalidate the outcomes of impact studies  (e.g., for hydrological model \cite{Grouillet2015SensitivityAO}).  (See Stainforth et al.\cite{Stainforth2007ConfidenceUA} for a detailed discussion on reasons for these biases.)

Climate scientists have developed a suite of tools termed Model Output Statistics (MOS) methods (defined in Maraun and Widmann, 2018, p.5 \cite{maraun_widmann_2018}) that correct these biases and provide output statistics that more closely match when compared to observations. Bias correction (BC) is a subset of MOS methods. However, the term BC is used differently in different communities (Maraun and Widmann, 2018, p.171 \cite{maraun_widmann_2018}). We use BC as synonymous with homogeneous (same physical variable) MOS with or without a spatial resolution mismatch. 

Most BC methods are univariate, correcting for a single one-dimensional physical variable in a single location. For example, the summer daily mean air temperatures in London \cite{Ho2012}. The most popular univariate methods are: mean correction \cite{Xu}, variance correction \cite{Boberg2012OverestimationOM, Ho2012, esd-11-1033-2020, Hawkins2013CalibrationAB, Iizumi2011ProbabilisticEO, Eden2012SkillCA, Schmidli2006DownscalingFG} and empirical quantile mapping (EQM) along with its variations \cite{Ho2012, Li2010BiasCO, SalviStatisticalDA, Wood2002LongrangeEH, Hagemann2011ImpactOA, Themel2011EmpiricalstatisticalDA, Turco2017BiasCA, Dqu2007FrequencyOP, vrac-2012, Kallache2010NonstationaryPD}. Additionally, there are multivariate BC approaches. They correct properties of multidimensional distributions \cite{robin-2019, piani-2012, cannon-2017, Dekens2017MultivariateDC, r2d2}) such as the joint distribution of daily temperatures and rainfall at multiple locations.  

However, BC methods currently struggle to adjust for temporal biases. As Vrac et al. \cite{R2D2-v2} explain, univariate methods cannot correct biases relating to temporal dependencies. Because they discard temporal information (see a simple illustration of the argument in Figure \ref{gpt_time_collapse}). This issue can be exacerbated by a spatial resolution mismatch between the reference and climate model data since univariate BC methods will struggle to reproduce the grid-scale temporal variability. For example, in the EQM method, a future projection value will be large if the climate model value is large, regardless of the specific observed behaviour at the corresponding grid-point. Additionally, most multivariate BC approaches are not designed to correct for statistics relying on the dependence between consecutive time points. Thus, they also struggle to correct temporal biases, as pointed out by Vrac et al. \cite{R2D2-v2} and shown by Francois et al.\cite{Franois2020MultivariateBC}.

There have been attempts to create BC methods that explicitly capture temporal properties. Vrac et al. developed a method \cite{Vrac2015MultivariateIntervariableSA} ('EC-BC') where the ranks of the climate simulations are shuffled to match those of observations. However, this method implies that the ranks of a future time series are the same as those in the present period. In this study, we show that its estimation of heatwave duration can be inaccurate when the assumption does not hold. Another method, by Mehrotra et al. \cite{Mehrotra2019ARA} ('3D-BC'), extends the Vrac et al. method to include 1-lag auto-correlation. This method has the advantage that it can capture the correlation of two consecutive readings by construction. However, extending it to higher-order correlations is not trivial. Vrac and Thao \cite{R2D2-v2} introduced a rank-resampling-based BC method ('R2-D2v2'), which aims to adjust multivariate properties while constraining the temporal properties to be similar to those of the references. However, it cannot be used in a homogeneous MOS setup since it requires an additional pivot variable assumed to be correctly ranked. More recently, Robin et al.\cite{Robin2021IsTA} proposed the 'TSMBC' method, which relies on using the multivariate BC method 'dOTC' \cite{Robin2019DOTC} in a particular configuration. It enables the adjustment of temporal dependencies by incorporating lagged time series as additional variables to be corrected. However, the authors show that limiting the temporal constraints to only a few timestamps is required. Consequently, it is limited in estimating different statistics, such as long heatwvaes. Despite these gaps, known impact studies about future heatwaves rely on univariate BC methods or the raw outputs of climate models \cite{Marx2021HeatWA, summers2022}. 

The goal of this paper is to introduce a BC method that can correct climate models and more accurately estimate future climatic statistics with long-term temporal properties, such as heatwaves. 

We find it helpful to offer a new BC perspective to achieve this goal. We cast the classic BC — typically treated as an algorithmic procedure mapping between histograms — as a probability model with a specific time index and without independence assumptions. Having a time-indexed target random variable as part of the regression task does not mean we can precisely specify the value obtained in Tokyo, Japan, on January 1st, 2050. However, there are still multiple benefits, such as: (a) we can obtain the distribution at that specific time-point, and this distribution can represent the typical behaviour for that month, season or decade, and (b) we can account for temporal correlations --- what happened in day $t$ can affect the possible values at day $t+1$. (For a detailed discussion on motivation for our probabilistic perspective, see Appendix \ref{motivation}.)

We are aware that our approach is unconventional, mainly since climate models and observations are not in temporal synchrony.  However, by including past observations and (past, future) climate model outputs and choosing a flexible machine learning (ML) architecture, we can learn the distribution at a specific time-point and temporal correlations despite the temporal asynchrony. To implement our BC method, we have devised a novel ML methodology based on the Taylorformer attention model \cite{Nivron2023TaylorformerPP, Vaswani2017}.

To put our method in context, we note that other methods model temporal properties explicitly. However, these methods are not strictly homogeneous MOS methods like ours. These methods include some form of Perfect Prognosis, i.e., they link statistically large-scale phenomena with local-scale phenomena in the observed data and apply the learned link to the climate model. Hence, we do not detail them further or use them as benchmarks. (For an example, see a conditional weather generator by Chandler \cite{CHANDLER}.)

To showcase the potential of our methodology, we test it on two case studies. We aim to correct the heatwave duration statistics in Abuja, Nigeria and Tokyo, Japan, which have very different climate characteristics and distribution shifts (see Appendix \ref{appendix_shift}). We use the simplified heatwave duration definition for our case studies (as a proof of concept): at least three consecutive days above a chosen absolute value in $^\circ C$. Our method performs accurately on the standard BC evaluations, such as QQ plots. Further, our model consistently outperforms the alternatives on the heatwave duration statistic for different thresholds during 20 years (1989-2008). For the $22 ^\circ$C threshold in Tokyo, our model misestimated the observed heatwaves by just $1\%$ (averaged over 32 initial-condition runs). In contrast, the next-best BC model, ’TSMBC’, misestimated the observed heatwaves by $8\%$ (averaged over 32 initial-condition runs). For the $24 ^\circ$C threshold in Abuja, our model misestimated the observed heatwaves by $0.5\%$, whereas the second-best BC model, in this case the mean-shift, misestimated the observed heatwaves by $9\%$. Lastly, our model outperforms all the other methods on the log-likelihood score. We believe these results support a further extension of this study to check its generalisation potential. (See results in the experiment section \ref{experiments_section} and extended results in Appendix \ref{appendix_experiments}.)

\section{Reference and model data}
\label{Refer}
In any BC task, data from a climate model is matched to real-world observations (or a proxy of observations, namely a reanalysis product) to correct biases. Because GCMs are typically coarser in resolution than observations or reanalysis reference data, the BC task is often joined by a downscaling of resolution.

We follow the same approach. For the climate model data, we chose the 32 initial condition runs from the climate model of the Institut Pierre-Simon Lapace (IPSL) \cite{Lurton-ipsl} \cite{Boucher-ipsl} \cite{Hourdin-ipsl} under the sixth International Coupled Model Intercomparison Project (CMIP6) \cite{Eyring2016} historical experiment. The data is obtained at a 1-day frequency for the variable 'tasmax', representing the maximum temperature 2 meters above the surface. The IPSL model is run at a 250km nominal resolution and is not re-gridded. For the case studies, we take the closest geographical point to Abuja, Nigeria, and the closest point to Tokyo, Japan, from 1948-2008. 

For the observational reference data, we use two different Reanalysis datasets:
\begin{itemize}
    \item For Abuja, Nigeria, we use the NCEP-NCAR Reanalysis 1 \cite{Kalnay1996TheN4} of the National Oceanic and Atmospheric Administration (NOAA). The data is obtained in six-hourly daily measurements for the variable 'tmax', the maximum temperature 2 meters above the surface, and is averaged to daily simulations. The resolution is 2.5 degrees globally in both the north-south and east-west directions and is not regridded. We chose the closest geographical point to Abuja, Nigeria. 
    \item For Tokyo, Japan, we use the ERA5 reanalysis \cite{era5} from the European Centre for Medium-Range Weather Forecasts (ECMWF). The data is obtained daily for the variable 'mx2t', the maximum temperature 2 meters above the surface. The resolution is 0.25 degrees globally in both the north-south and east-west directions and is not regridded. We chose the closest geographical point to Tokyo, Japan.
\end{itemize}

\noindent We use two different reanalysis datasets, NCEP-NCAR and ERA5, to show that the methodology is not limited to one spatial resolution configuration. We hypothesize that the model will also work for many different configurations of (climate model, reanalysis) pairs. 

\section{Methodology}
To start, let us first describe the big picture of our method. Our goal in this study is to estimate potential continuations of observed values from 1989 to 2008, given past observations and (past, future) climate model outputs. To do so, we (1) generate a custom dataset comprised of multiple (climate model, observation) pairs with varying time-series lengths from the reference and climate model data, (2) train a probability model with an explicit likelihood function \footnote{a likelihood function is a measure of how likely a given set of parameters would result in the observed data. Some models, like hidden Markov models, do not have a closed-form likelihood function.} on the custom data, and (3) generate entire time-series samples from the trained probability model to continue the reference data to the future (we call this step the "Inference procedure", as is common in the ML literature). We use the generated samples to estimate different statistics. Here, we concentrate on estimating the distribution of 'heatwave duration'.

In the following, we will describe our approach in greater detail, starting in reverse order to explain the motivation of 
our end goal. We first detail the generation process using a trained probability model (step 3) before explaining how to train it (steps 1 and 2). To be more precise, we introduce mathematical notation. Throughout the text, we will use uppercase letters to denote random variables and lowercase letters to denote specific values. Let $O_i$ be a random variable representing real-world reading of maximum daily temperatures at time $t_i$, and $g_i$, the GCM output of maximum daily temperatures at time $t_i$. Bolded variables are vectors, and the notation 
$i:j$ denotes the set of all integer indices from $h$ through $j$, inclusive. 

\subsection{Inference procedure} \label{infer_section}
Using the notation above, our goal (step 3 above) is to estimate $\textbf{O}^{\star} = \left(O_1^{\star}, \dots O_m^{\star} \right)$ at some time-points $t^{\star}_1, \dots t^{\star}_m$ given GCM outputs $\textbf{g} = (g_1, \dots, g_\ell)$ at time-points $t'_1, \dots, t'_\ell$ and observed real-world readings $\textbf{o} = (o_1, \dots, o_n)$ at $t_1, \dots, t_n$. We do not assume synchronicity: the climate model time-points $t'_1, \dots, t'_\ell$ may in general be different to times $t_1, \dots t_n, t^{\star}_1, \dots t^{\star}_m$.

Since we start in reverse order, we assume we already have trained some probabilistic model that can generate samples of $\textbf{O}^{\star}$ given $\textbf{g}$ and $\textbf{o}$ (and time-points). We will additionally suppose that this probability model admits an explicit likelihood function, call it $P_\theta\left(\textbf{O}^{\star} \middle| \textbf{g}, \textbf{o}\right)$, where $\theta$ represents the learnable parameters. We call $\textbf{o}$ and \textbf{g} the conditioning set.

Generating samples of  $\textbf{O}^{\star}$ given $\textbf{g}$ and $\textbf{o}$ is a useful goal for estimating many different statistics. Because, for example, if we sample many time-series, we can estimate the marginal distribution of a specific time-point, such as $O_1^*$. Separately, we can estimate the average of $\textbf{O}^{\star}$. Furthermore, for each generated time-series we can estimate the number of consecutive days above a chosen threshold ('heatwave duration').  

\begin{figure}[t]%
\FIG{\includegraphics[width=1\textwidth]{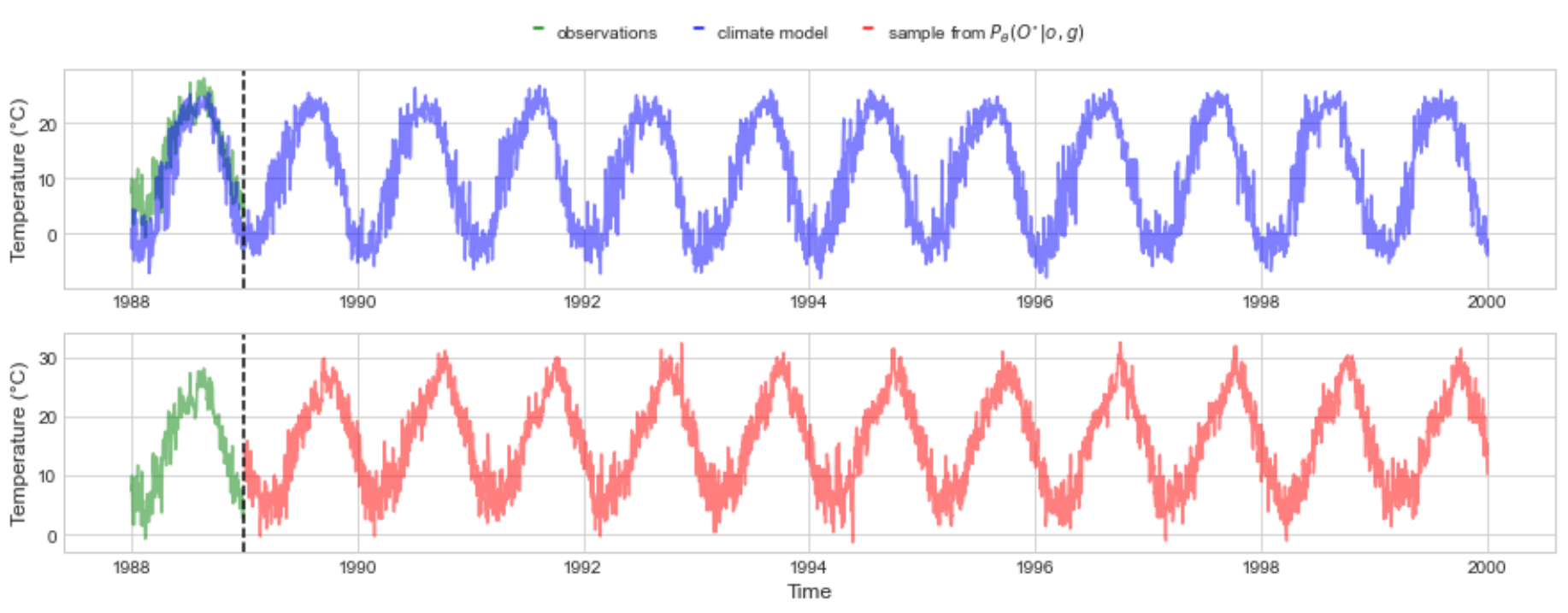}}
{\caption[]{\textbf{Inference Task for Estimating Future Observations:} The top panel outlines the available data to us in order to estimate the potential continuation of observed values post-1989 (dashed vertical line), based on historical observations (green line) and both past and future climate model outputs (blue line). The bottom panel displays a red line representing one possible continuation, sampled from the forecasting model which estimates $P_\theta\left(\textbf{O}^{\star} \middle| \textbf{o}, \textbf{g}\right)$}   
\label{inference_fig}}
\end{figure}

Figure \ref{inference_fig} shows in the top panel the data available to us at the start of the inference task, and the bottom panel shows an example of a generated sample time-series. 

However, training a probability model to generate sensible time-series—ones that yield accurate statistics compared to observed hold-out set data—is difficult because climate models and observations are not synchronised temporally (Maraun and Widmann, 2018, \cite{maraun_widmann_2018}), and we do not know the exact form of asynchrony.

The following section provides the blueprint for training a chosen probability model. After which, in section \ref{Taylorformer_section}, we discuss the specific architecture choice we have made in this study.    

\subsection{Training procedure} \label{training_section}
Assume we start with a training dataset that consists of real-world readings $o_1, \dots, o_n$ at $t_1, \dots t_n$ and climate model outputs $g_1, \dots g_\ell$ at $t'_1, \dots t'_\ell$. We emphasise in the notation that the climate model time-points do not have to be synchronised with the real-world reading time-points. However, we will assume they are synchronised for the readability of the data generation and training algorithms.  

As described in section \ref{infer_section}, our goal is to train a probability model that can generate time-series samples. One such probability model is the Taylorformer \cite{Nivron2023TaylorformerPP}, described in section \ref{Taylorformer_section}, which we use for our evaluation.

\begin{figure}[ht]%
\FIG{\includegraphics[width=1\textwidth]{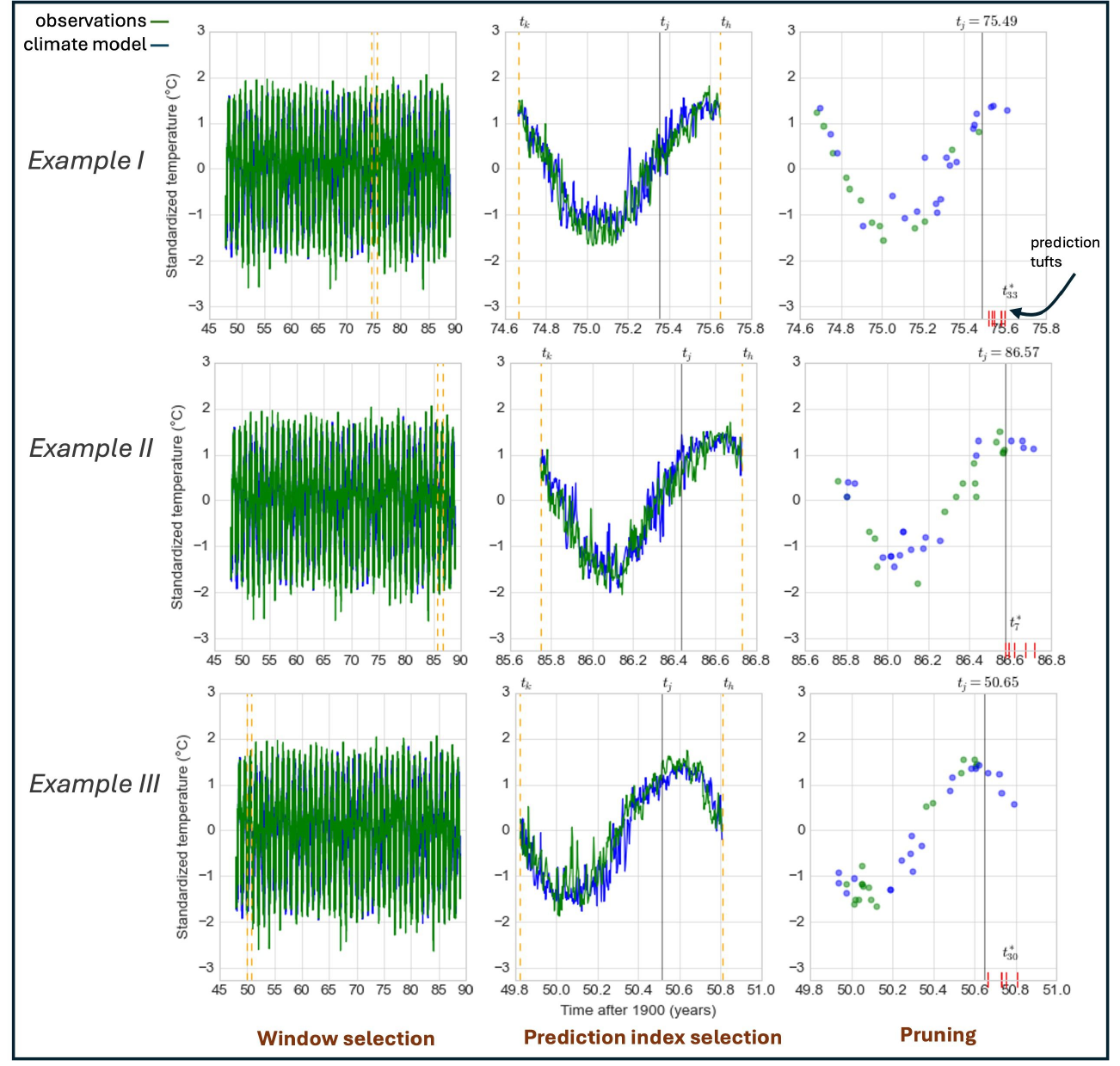}}
{\caption[]{\textbf{ Construction of Training Data for Tokyo, Japan: }Column I displays full sequences from 1948 to 1988 for climate models (blue) and observations (green). Dashed orange lines indicate the selected slice for each data row, a process termed "Window Selection." Column II zooms into the selected window, featuring a vertical black line at a randomly selected "prediction index" ($t_j$), from which we aim to estimate the observations until 
$t_h$. Column III illustrates the "Pruning" operation, where time points before and after $t_j$ are randomly selected from both the climate model and observations. The observed values to be estimated are concealed, and the chosen time points, called "prediction tufts" are highlighted with red tufts. This column adapts the data for use in ML sequential models. Note that we have dropped the example index $i$ for readability, but $k, j, h$ change from one example to the next.}
\vspace{-10pt}
\label{training_construction}}
\end{figure}

In Algorithm \ref{train_batch_generation}, we describe the precise steps to generate a training batch. The motivation for the procedure will become apparent in section \ref{Taylorformer_section}. In loose language, an element $i$ in the training batch is a pair of (climate model, observations) time-series. The element is constructed by (i) randomly picking a window of 60 to 360 aligned (in-time) points both from observations and GCM outputs. We denote the resulting climate model time-series as $\textbf{g}^{i}$. Then (ii) randomly picking a time-point $t_j$. This time-point divides the observed time-series, such that all points in the observations after $t_j$ become the predictands, $\textbf{O}^{\star, i}$ and all points before $t_j$ are given in $\textbf{o}^{i}$. The dataset creation procedure is schematically shown in Figure \ref{training_construction}. The dashed orange lines represent the chosen window, while the black vertical line shows $t_j$.  

After constructing a training batch, we can use Algorithm \ref{train_algo} to optimise the parameters $\theta$. Note that we have set the elements $\textbf{O}^{\star, i}$, $\textbf{o}^{i}$ and $\textbf{g}^{i}$ when constructing the batch in Algorithm \ref{train_batch_generation} and we have assumed we have an explicit likelihood function. So, we just need to plug the elements into the likelihood function and perform the optimisation. (We call the likelihood score the "Training objective".)       
\begin{algorithm}
\caption{Procedure to generate a random training batch}
\label{train_batch_generation}
\begin{algorithmic}[1]
\State \textbf{Input:} batch size $B$, observed sequence $o_1, \dots, o_n$, climate model sequence $g_1, \dots, g_n$ (and associated time-points)
\For{$i = 1$ \textbf{to} $B$}
    \State \LeftComment{"Time window boundary selection" for the training sample}
    \State Pick $k$ uniformly at random from $\{1, \ldots, n-360\}$ \Comment{left boundary}
    \State Pick $h$ uniformly at random from   $\{k+60, \ldots, k+360\}$ \Comment{right boundary}
    \State \LeftComment{"Prediction index selection" within the window}
    \State Pick $j$ uniformly at random from $\{k+5, \ldots, h-5\}$
    \State \LeftComment{Construct training sample $i$}
    \State \quad Set  $C_i = \{v \mid t_k \leq t_v \leq t_j\}$
    \Comment{Context indices}
    \State \quad Set  $A_i = \{v \mid t_j < t_v \leq t_h\}$
    \Comment{Prediction indices}
    \State \quad Set $\textbf{O}^{\star, i} = \left[o_v \mid v \in A_i\right]$
     \Comment{Elements to be predicted}
    \State
    \LeftComment{ observations and climate model elements for conditioning set}
    \State \quad Set $\textbf{o}^{i} = \left[o_v \mid v \in C_i \right]$
    \State \quad Set $\textbf{g}^{i} = \left[g_v \mid v \in C_i \cup A_i \right]$
\EndFor
\end{algorithmic}
\end{algorithm}

\begin{algorithm}
\caption{Training algorithm}
\label{train_algo}
\begin{algorithmic}[1]
\While{\textbf{true}} \Comment{break loop when objective is bigger than some choosen threshold}
    \State Generate a random batch of time-series pairs of size $B$ \Comment{using Algorithm \ref{train_batch_generation}}
    \State Perform gradient descent to maximise the objective:
    \State \quad $\sum_{i=1}^{B} \log P_\theta \left(\textbf{O}^{\star, i} \mid \textbf{g}^{i}, \textbf{o}^{i}\right)$ \Comment{"Training objective"}
\EndWhile
\end{algorithmic}
\end{algorithm}

The probability model we use in Algorithm \ref{train_algo} can be theoretically any model that can generate samples from  $P_\theta\left( \textbf{O}^{\star}\middle| \textbf{o}, \textbf{g}\right)$, has an explicit likelihood form and uses the constructed training data to do so. For example, a linear regression model can be used as the forecasting model, but since it assumes by construction independence among the elements of $\textbf{O}^{\star}$ it will, most likely, produce a bad fit.

The inference procedure in Section \ref{infer_section} and the training procedure in Section \ref{training_section} show the general purpose framework. We now discuss the specific choice of probability model we have made in this study.

\subsection{Taylorformer as the probability model} \label{Taylorformer_section}
In the present study, we have chosen the Taylorformer \cite{Nivron2023TaylorformerPP}, which is a sequential ML model adapted from the Transformer attention model by Vaswani \cite{Vaswani2017} to real-valued targets on a continuum. In Appendix \ref{Taylorformer_appendix}
we provide a detailed explanation of the Taylorformer. The explanation starts with the Transformer decoder for language modelling and the needed adaptations to reach the Taylorformer for BC. The motivation for choosing Taylorformer is its ability (not shown here) to learn different types of biases such as mean bias, variance inflation or deflation, and temporal a-synchronicity. The Taylorformer learns these biases from data without specifying a-priori the parametric form of the bias, which is exactly what we need since we do not know the time asynchronicity between GCMs and observations. Furthermore, the Taylorformer and other attention-based models can learn changing "rules" for temporal synchronicities. Their success in dealing with "temporal asynchronicity" is shown when translating a sentence from one language to another. For example, the subject of a sentence may change its position from one language to another. We now explain the adaptations to the Taylorformer needed for the BC task.

\paragraph{Training objective} 
Since the Taylorformer is a sequential model, it is a standard practice to decompose the "Training objective" in Algorithm \ref{train_algo} according to the probability chain rule. \mbox{We write the training objective as}
\begin{align}
\operatorname{maximise}_{\theta} \operatorname{log} P_{\theta}(\textbf{O}^{\star, i} \mid \textbf{o}^{i}, \textbf{g}^{i}) &= \\\operatorname{maximise}_{\theta} \operatorname{Log} 
    &P_{\theta}\left(O^{i}_{h} \middle| O^{i}_{{j:{h-1}}}, \textbf{o}^{i}, \textbf{g}^{i}\right) \dots P_{\theta}\left(O^{i}_{{j+1}}\middle| O^{i}_{j}, \textbf{o}^{i}, \textbf{g}^{i}\right) P_{\theta}\left(O^{i}_{j}\middle| \textbf{o}^{i}, \textbf{g}^{i}\right) \notag
\end{align}

\noindent We model each term in the decomposition using a Normal distribution. Specifically, for any integer index \( s \) where s is bigger than the "prediction index" $j$, we define:
\begin{equation}
\label{prob_model}
    O^{i}_s \mid \textit{inp}^{i}_{s} \sim \operatorname{Normal}\left(f_\theta\left(\textit{inp}^{i}_{s}\right), g_\theta\left(\textit{inp}^{i}_{s}\right)\right)
\end{equation}

\noindent where we define $\textit{inp}^{i}_{s} = \left(O^{i}_{j}, \dots O^{i}_{s-1}, \textbf{o}^{i}, \textbf{g}^{i}\right)$. In our model $f_\theta\left(\textit{inp}^{i}_{s}\right)$  represents the mean value of the normally distributed $O^{i}_s$ and  $g_\theta\left(\textit{inp}^{i}_{s}\right)$ its variance. Therefore, we get a distribution for $O^{i}_s$, i.e., stochastic outputs. Here, the functions $f_\theta$ and $g_\theta$ are defined by the Taylorformer architecture.  

 \paragraph{Training dataset}
Practically, to train our dataset towards our inference goal, we must do more than feed in the entire time-series of the pair (observations, climate model) from 1948 to 1988. Feeding in the entire time-series would be fine for classical models such as ARIMA \cite{box1970time} or GPs. Differently, sequential deep learning models, such as the Taylorformer or a Recurrent Neural Network \cite{rumelhart1986learning}, cannot learn from one sequence since this would enable only one 'gradient descent pass', and little learning would occur. For that reason, in the batch generation Algorithm  (\ref{train_batch_generation}), we break down the pair of (climate model, observations) into multiple pairs, as shown in the first two columns of Figure \ref{training_construction}. 

We have to address two possible issues with our newly constructed sequences. (1) We want to allow pairs to have differing lengths, as it is unclear a-priori which length is optimal; and (2) For a deep learning sequential model, we have to heed that it does not learn the rule \say{Tomorrow will be the same as today} as pointed out by Nivron et al. \cite{Nivron2023TaylorformerPP}. The Taylorformer can address both points if we feed it in with an appropriate dataset, as explained next. 

To address point (1), the Taylorformer \cite{Nivron2023TaylorformerPP} architecture can receive as input arbitrary length sequences by construction. Therefore, we simply need to feed in sequences of differing lengths as part of our training procedure (see "Time window selection" in Algorithm \ref{train_batch_generation}).

To address point (2), we employ the following "Pruning"  mechanism: after we set $\textbf{O}^{\star, i}$ and $\textbf{o}^{i}, \textbf{g}^{i}$ in Algorithm \ref{train_batch_generation}, we randomly prune elements from them. Let us denote the results of the operation as $\operatorname{Prune}(\textbf{O}^{\star, i})$, $\operatorname{Prune}(\textbf{o}^{i})$ and $\operatorname{Prune}(\textbf{g}^{i})$. We call the time-points associated with $\operatorname{Prune}(\textbf{O}^{\star, i})$ the "prediction tufts".
We show the "Pruning" operation and the "prediction tufts" in the third (rightmost) column of Figure \ref{training_construction}. Intuitively, if the model does not receive as input the value at the previous time-point at each iteration, it must search for a different 'rule' for correction instead of copying yesterday's value. Further, we limit the maximum sequence length to 360 points, which is chosen to address the high computational load of attention-based models such as the Taylorformer.

So we can now refine our training objective, incorporating the solutions to the two points above. The optimisation becomes 
\begin{equation}
\operatorname{maximise}_{\theta} \operatorname{Log} 
P_\theta\left(\operatorname{Prune}(\textbf{O}^{\star, i}) \middle| \operatorname{Prune}(\textbf{o}^{i}), \operatorname{Prune}(\textbf{g}^{i})\right)
\end{equation}
\noindent which we again decompose according to the probability chain rule. 
(Note that the training dataset could be constructed in different ways. We discuss other possibilities in Appendix \ref{discussion_slice_infer}.)

\paragraph{Multiple initial condition runs}
Our source data consists of one climate model ('IPSL') that was initialised 32 times to generate 32 different time-series. To include all the runs in our model, we add a "random run selection" step as the first command in the for loop in Algorithm \ref{train_batch_generation}.         

\begin{figure}[ht]%
\centering
\FIG{\includegraphics[width=0.5\textwidth]{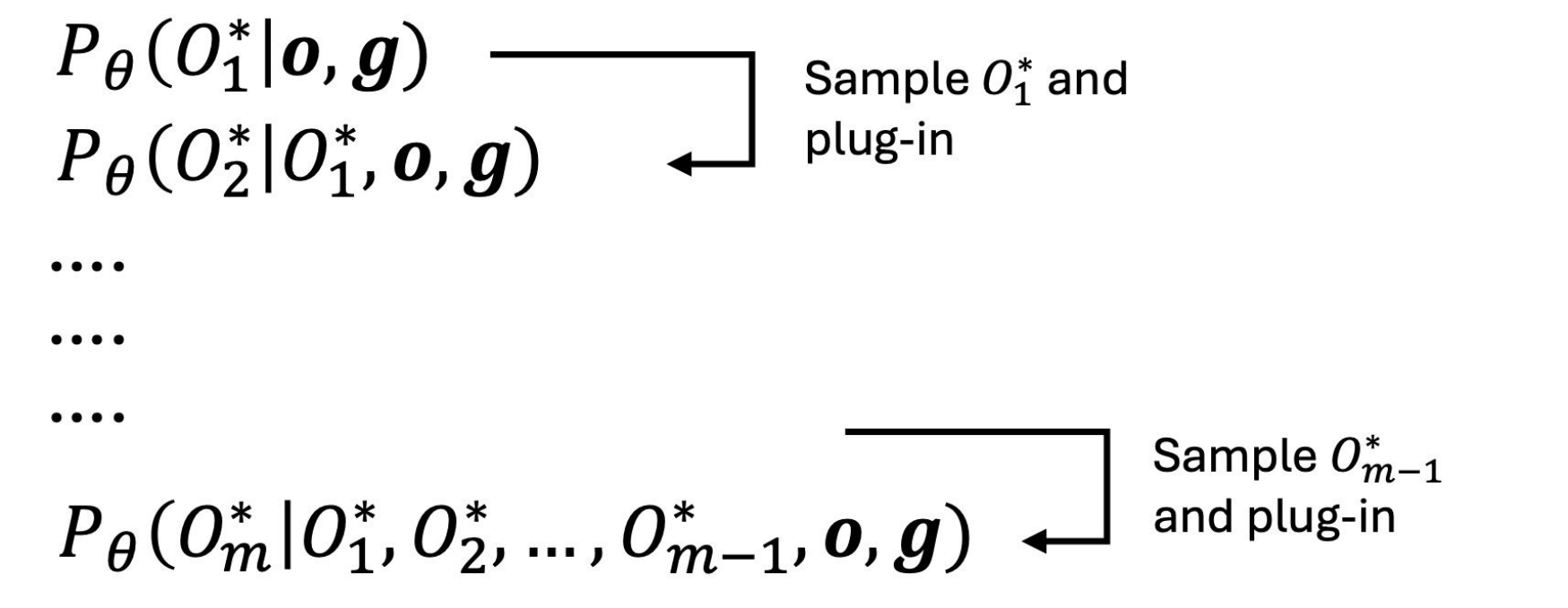}}
{\caption[]{Illustration of the inference procedure. Given a trained Taylorformer model with parameters $\theta$, we generate a time-series sequentially. In the top row, we generate a value $o_1^{\star}$ which will be then plugged into the condition set in the second row and so forth}   
\label{infer_eq_fig}}
\end{figure}

\paragraph{Inference}    
In ML, we first train a model to learn the parameters of a network. Once the parameters are learned using an objective (here, maximum log-likelihood), they are fixed, and we can use the network to get our desired readouts. This is called the "Inference stage". Our chosen Taylorformer model operates using sequential generation. In each step, we sample a value from a conditional distribution and plug the result into the condition set of the following conditional distribution. Figure \ref{infer_eq_fig} shows an illustration of the procedure. The bottom plot of Figure \ref{inference_fig} shows an example of a fully sampled time-series.  

The inference differs from the training procedure in which we attach the observed value (as opposed to a sampled value) to the new condition set. This is the standard way to train a sequential model using maximum likelihood (also known as 'teacher forcing' \cite{williams1989learning}).   

Due to computational constraints, we do not include in the conditioning set, $\textbf{o}, \textbf{g}$, all the observed data and (past, future) GCM data. Instead, we feed in the following sequence: $\textbf{o}$ contains the last 60 days of 1988. $\textbf{g}$ contains the last 60 days of 1988 and the next 120 days of 1989 from the GCM outputs. Then, given $\textbf{o}$ and $\textbf{g}$ we produce the first estimated observation of 1989, $O^{\star}_1$. Next, we attach the sampled value to our observations and shift all the processes by one day to the right so that when sampling the second day of 1989, $O^{\star}_2$, we still have a condition set of the same size. We iterate the process until we have all days up to 2008. (Note that the inference procedure could be done in different ways. We discuss other possibilities in Appendix \ref{discussion_slice_infer}.)

\section{Experiments} \label{experiments_section}
To test our model, we do the following experiment: First, we bias-correct daily maximum temperatures for Abuja, Nigeria, and separately for Tokyo, Japan, for 1989-2008 based on the years 1948-1988. Second, we calculate the heatwave duration distribution, our primary interest, from the bias-corrected data. 

A primary interest in using BC methods is to estimate the climate in the future. Here, we use 1989-2008 as our unseen 'future'. Both Tokyo and Abuja have experienced a distribution shift (see Appendix \ref{appendix_shift}) in maximum daily temperatures between our training period, 1948-1988, and the unseen 'future' period, 1989-2008, and hence serve as a good testing ground for our method. 

An important question for BC methods is whether they preserve the trend exhibited by the climate model. Unlike mean shift, our method does not necessarily preserve the trend of the climate model per initial condition run. We show it in Appendix \ref{appendix_experiments} in Figures \ref{tokyo_5_year_avg_trend_fig} and \ref{abuja_nigeria_5_year_avg_trend_preserv_fig} for Tokyo and Abuja, respectively. The figures represent the 5-year averages from 1989-2008 for different initial condition runs for the climate model ('GCM'), reanalysis data ('ERA') and our Taylorformer model ('ML'). The figures also show that different initial condition runs from the climate model exhibit different trends, and in turn, those trends, in many cases, do not match the 'observed' (reanalysis) trend.   

\paragraph{Comparison}
We compare our model to the raw output from the GCM (IPSL) and to six BC methods: mean correction \cite{Xu}, mean and variance correction \cite{Ho2012}, EQM \cite{panofsky1968some}, EC-BC \cite{Vrac2015MultivariateIntervariableSA}, 3DBC \cite{Mehrotra2019ARA} and TSMBC \cite{Robin2021IsTA}. Note that due to the spatial resolution mismatch, the mean correction, mean and variance correction and EQM methods potentially misrepresent the local scale variability (see Maraun and Widmann, 2018, p.189-192   \cite{maraun_widmann_2018} ). Nevertheless, we also compare with them since these are very popular BC methods (and, in turn, used for heatwave assessments) and are practically used with a spatial mismatch on many occasions, for example, in Vrac and Friederichs  \cite{Vrac2015MultivariateIntervariableSA}. The full implementation details of the different BC methods are provided in Appendix \ref{implementation details}.

\subsection{Results}
\begin{figure}[ht]
    \centering
    \begin{minipage}{0.48\textwidth}
        \centering
        \includegraphics[width=\linewidth]{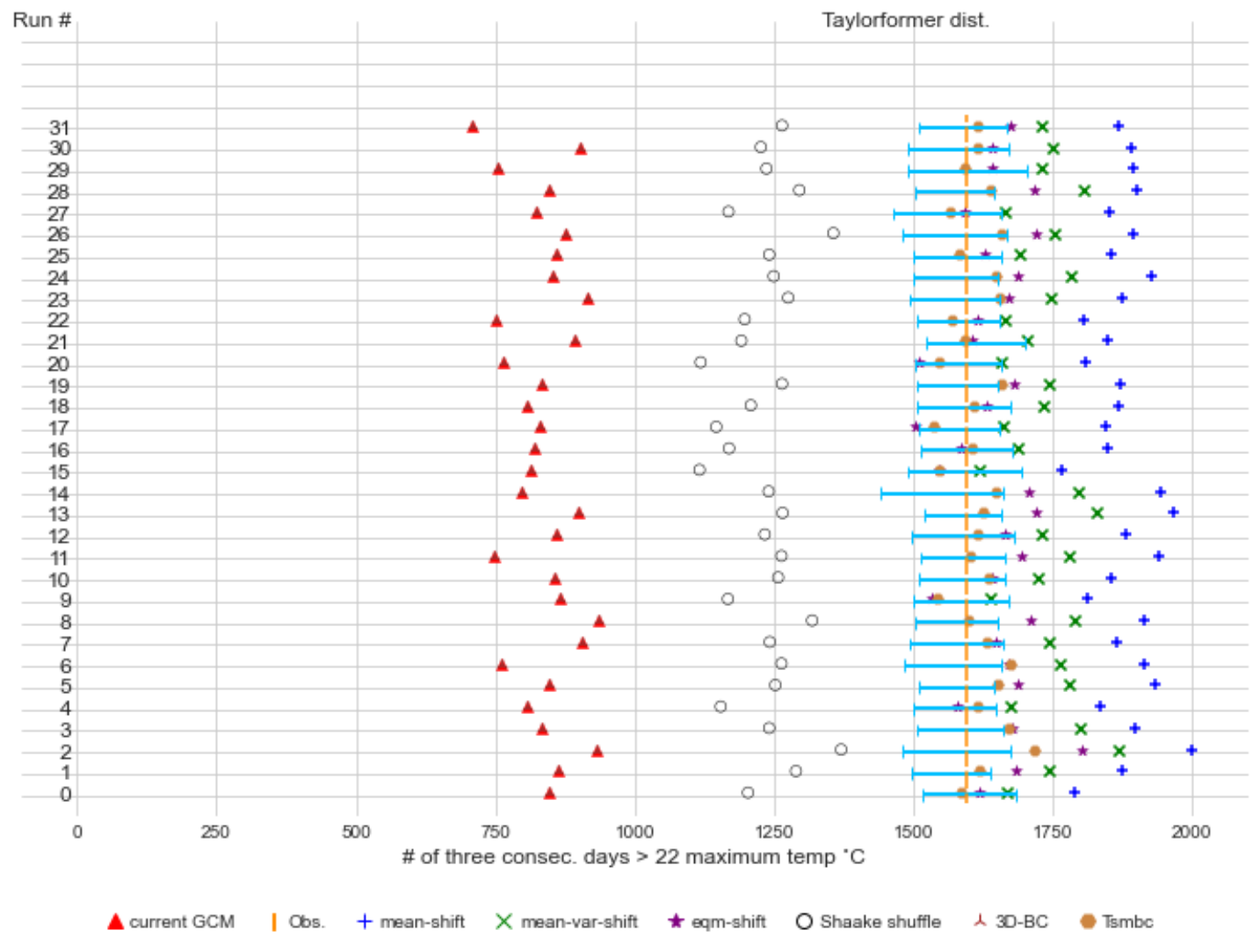}
        \caption[The raw outputs from the IPSL climate model underestimate...]{\textbf{Comparative Analysis of 'heatwave duration' Trends in Tokyo, Japan (1989-2008):} The number of periods featuring at least three consecutive days with temperatures exceeding 22$^\circ$C is shown. The IPSL climate model predictions are represented by red triangles, which generally underestimate the observations. Actual observations are indicated by a vertical orange line. The Taylorformer temporal BC are depicted using horizontal box plots, with whiskers indicating the 1st and 3rd quartiles. Markers for other BC methods are indicated at the bottom of the figure.}
        \label{tokyo_22_fig}
    \end{minipage}\hfill
    \begin{minipage}{0.48\textwidth}
        \centering
        \includegraphics[width=\linewidth]{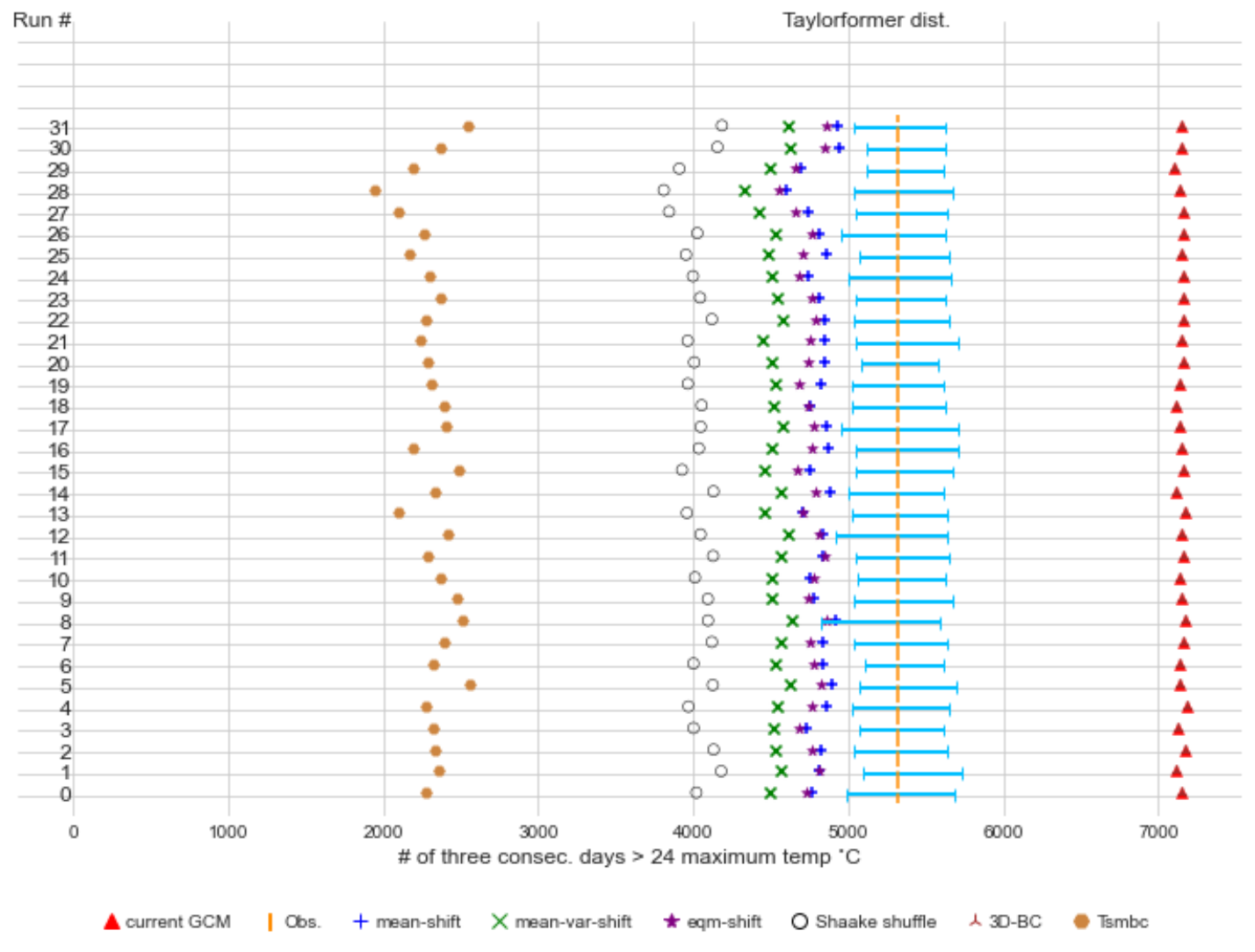}
        \caption[The raw outputs from the IPSL climate model overestimate...]{\textbf{Comparative Analysis of 'heatwave duration' Trends in Abuja, Nigeria (1989-2008):} The number of periods featuring at least three consecutive days with temperatures exceeding 24$^\circ$C is shown. The IPSL climate model predictions are represented by red triangles, which overestimate the observations. A vertical orange line indicates actual observations. The Taylor former temporal BC is depicted using horizontal box plots, with whiskers indicating the 1st and 3rd quartiles. Markers for other BC methods are indicated at the bottom of the figure.}
        \label{abuja_24_fig}
    \end{minipage}
\end{figure}

Figure \ref{tokyo_22_fig} shows the results for the 'heatwave duration' distribution for Tokyo, Japan using a threshold of 22$^\circ$C. The raw outputs from the IPSL GCM (red triangles) underestimate the number of instances of observed heatwaves (vertical orange line) by more than $100 \%$ for the period $1989-2008$.  Our novel Taylorformer temporal BC produces a much more accurate distribution (horizontal box-plots) per GCM initial condition run (0-31) for the number of heatwaves in the same period. The Taylorformer's average difference is $0.9\%$ to the observed number of heatwaves. Other BC models perform worse: $17.5\%, 8.5\%, 3.1\%, 22.7\%, 47.5\%, 1.1\%$ for mean-shift, mean and variance shift, EQM, EC-BC, 3D-BC and TSMBC respectively.

Figure \ref{abuja_24_fig} shows the results for the 'heatwave duration' distribution for Abuja, Nigeria, using a threshold of 24$^\circ$C. The raw outputs from the IPSL GCM (red triangles) overestimate the number of instances of observed heatwaves (vertical orange line) by more than $33 \%$ for the period $1989-2008$. Our novel Taylorformer temporal BC produces a much more accurate distribution (horizontal box-plots) per GCM initial condition run (0-31) for the number of heatwaves in the same period. The Taylorformer's average difference is $0.5\%$ to the observed number of heatwaves. Other BC models perform worse: $10\%, 15\%, 11\%, 24\%, 34\%, 56\%$ for mean-shift, mean and variance shift, EQM, EC-BC, 3D-BC and TSMBC, respectively.

For completeness, we provide QQ plots and auto-correlation plots in Appendix \ref{appendix_experiments}. 
From the QQ plots, we see that our model performs at least as well as the EQM. The auto-correlation plots are more challenging to interpret. We also provide the $\operatorname{MSE}$ and log-likelihood scores. Our method is more accurate on both. 
The log-likelihood on a hold-out set gives us a combined measure of skill and spread and is thus informative. In contrast, the $\operatorname{MSE}$ is not very informative, typically, for evaluating BC tasks on the resulting time series, since it does not provide us with a measure of the spread of the process. Nevertheless, since our method continues the observed time-series, the $\operatorname{MSE}$ may be informative on short time-scales (we leave the analysis of this point for future work).    

\section{Conclusions}
This paper offers a novel stochastic temporal BC method. We test it on the correction of daily maximum temperatures time series in Abuja, Nigeria, and Tokyo, Japan. It performs accurately on standard BC evaluations such as QQ plots and is also consistently more accurate than other BC methods on heatwave duration statistics. The method is constructed by first re-thinking BC as a time-indexed probability model and then adapting state-of-the-art attention ML models. Our model opens up the opportunity to do BC using a framework different from what is currently common, with potential accuracy gains regarding temporal properties and no apparent loss about the more usual distributional evaluations. Subsequently, these accuracy gains may have a large impact on policy-making decisions. Going forward, this study needs to be scaled up, considering multiple regions, more climate models and more climate statistics. This can be done, for example, by applying it to experiments within dedicated frameworks, such as the VALUE framework \cite{VALUE}.

\begin{appendix}

\section{Distribution Shift} \label{appendix_shift}
In the context of climate change, we are often interested to see if a specific location and a specific physical variable (e.g., daily maximum temperatures) has seen a change in its distribution. 

Here, we show that for selected months, the two chosen locations, Tokyo and Abuja, had experienced a shift in their distributions. It is apparent both in the observed record (ERA5) and in the climate model (IPSL), and their respective changes seem to be qualitatively different.    
\begin{figure}[H]
    \centering
    \begin{minipage}{0.48\textwidth}
        \centering
        \includegraphics[width=\linewidth]{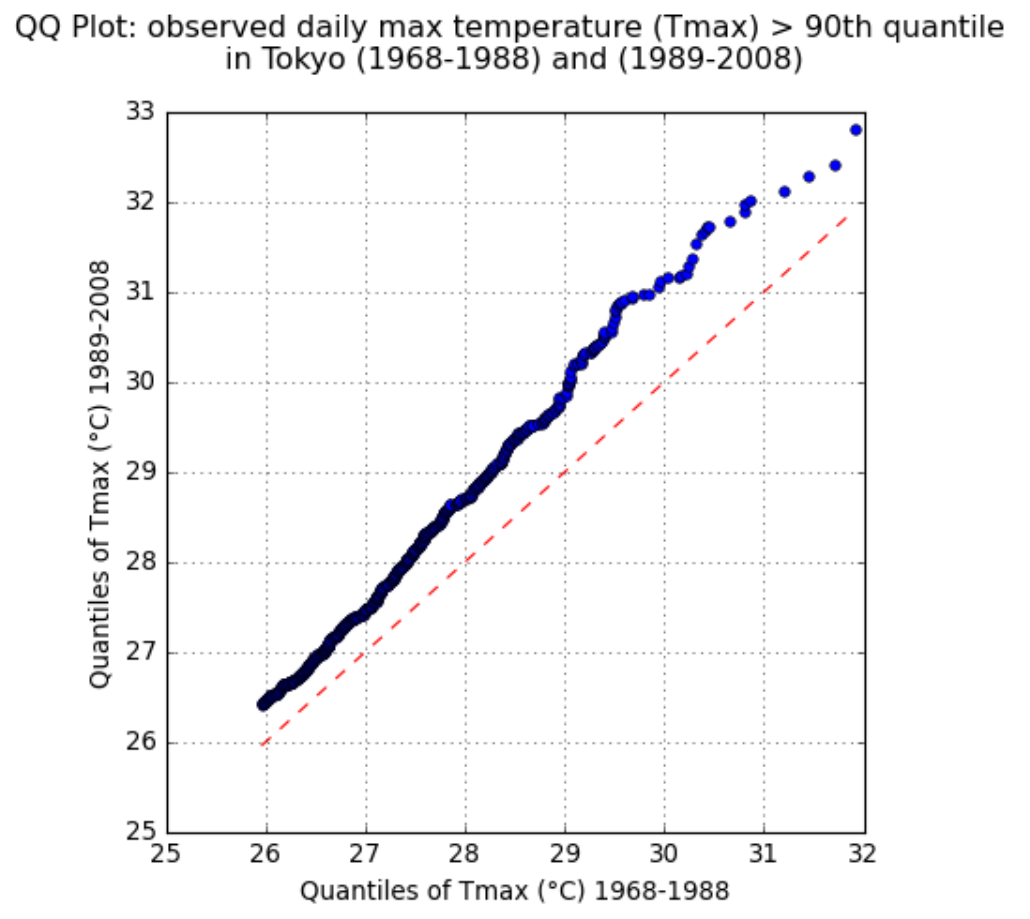}
        \caption[]{\textbf{Changes in the Distribution of Daily Maximum Temperatures Above the 90th Quantile in Tokyo:} This plot compares the distribution of temperatures between two periods: 1968-1988 (X-axis) and 1989-2008 (Y-axis)}
    \label{tokyo_qq_above_90_quantile}
    \end{minipage}\hfill
    \begin{minipage}{0.48\textwidth}
        \centering
        \includegraphics[width=\linewidth]{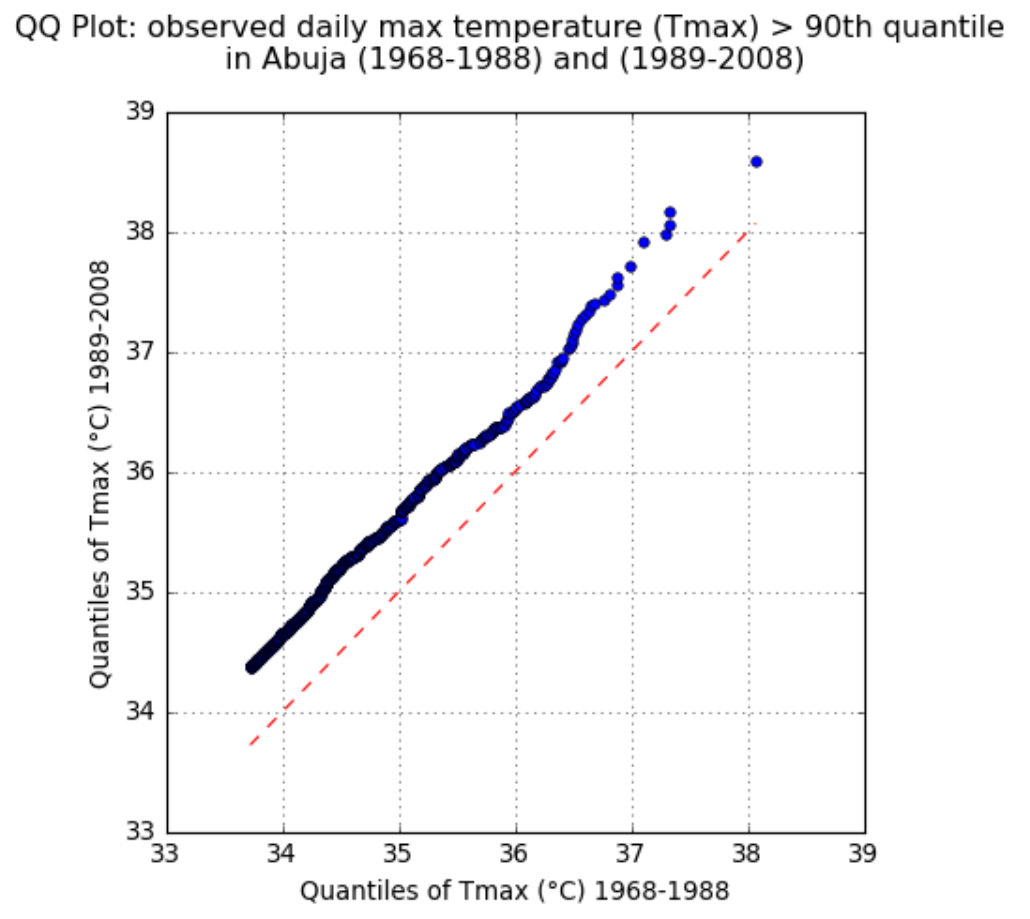}
        \caption[]{\textbf{Changes in the Distribution of Daily Maximum Temperatures Above the 90th Quantile in Abuja:} This plot compares the distribution of temperatures between two periods: 1968-1988 (X-axis) and 1989-2008 (Y-axis)}\label{abuja_qq_above_90_quantile}
    \end{minipage}
\end{figure}
\vspace{-10pt}

\begin{figure}[H]
    \centering
    \begin{minipage}{0.48\textwidth}
        \centering
        \includegraphics[width=\linewidth]{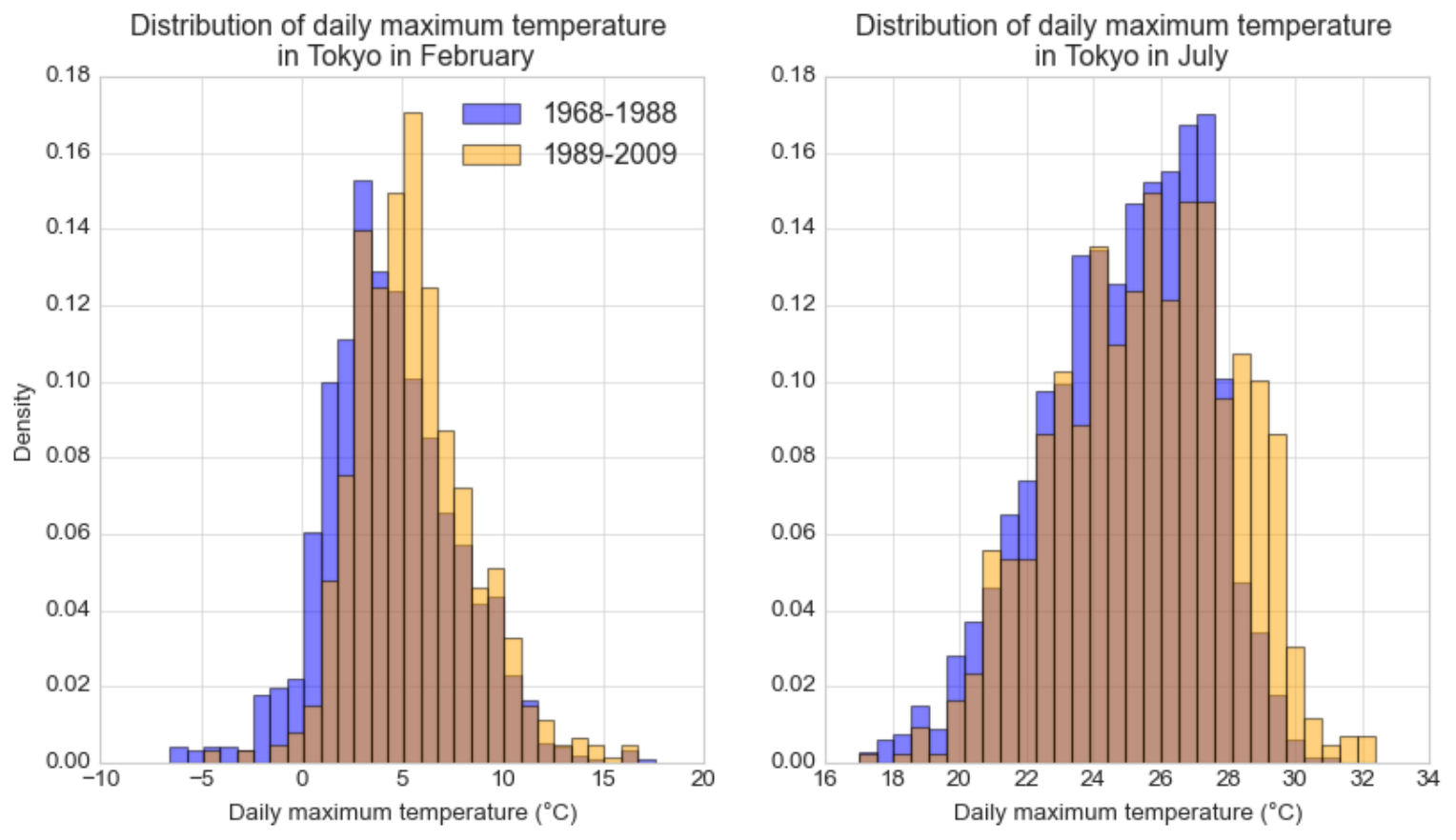}
        \caption[]{The distribution of daily observed (ERA5) maximum temperatures ($^\circ C$) in Tokyo, Japan had changed from 1968-1988 (blue histogram) to 1989-2009 (orange histogram) in February (left) and July (right).}
        \label{tokyo_obs_feb_july}
    \end{minipage}\hfill
    \begin{minipage}{0.48\textwidth}
        \centering
        \includegraphics[width=\linewidth]{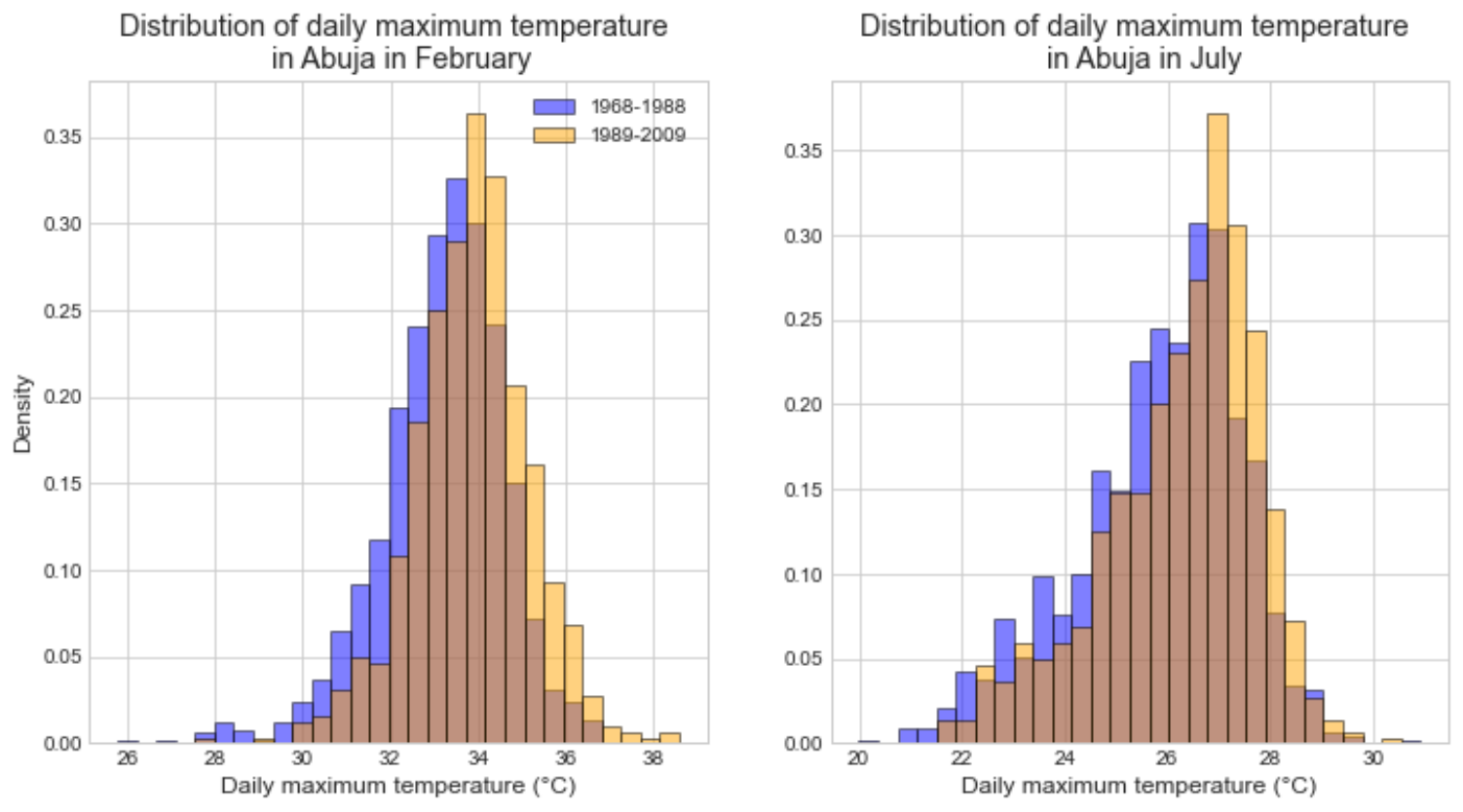}
        \caption[]{The distribution of daily observed (ERA5) maximum temperatures ($^\circ C$) in Abuja, Nigeria had changed from 1968-1988 (blue histogram) to 1989-2009 (orange histogram) in February (left) and July (right).}
        \label{abuja_feb_daily_temp}
    \end{minipage}
\end{figure}

\begin{figure}[ht]
    \centering
    \begin{minipage}{0.48\textwidth}
        \centering
        \includegraphics[width=\linewidth]{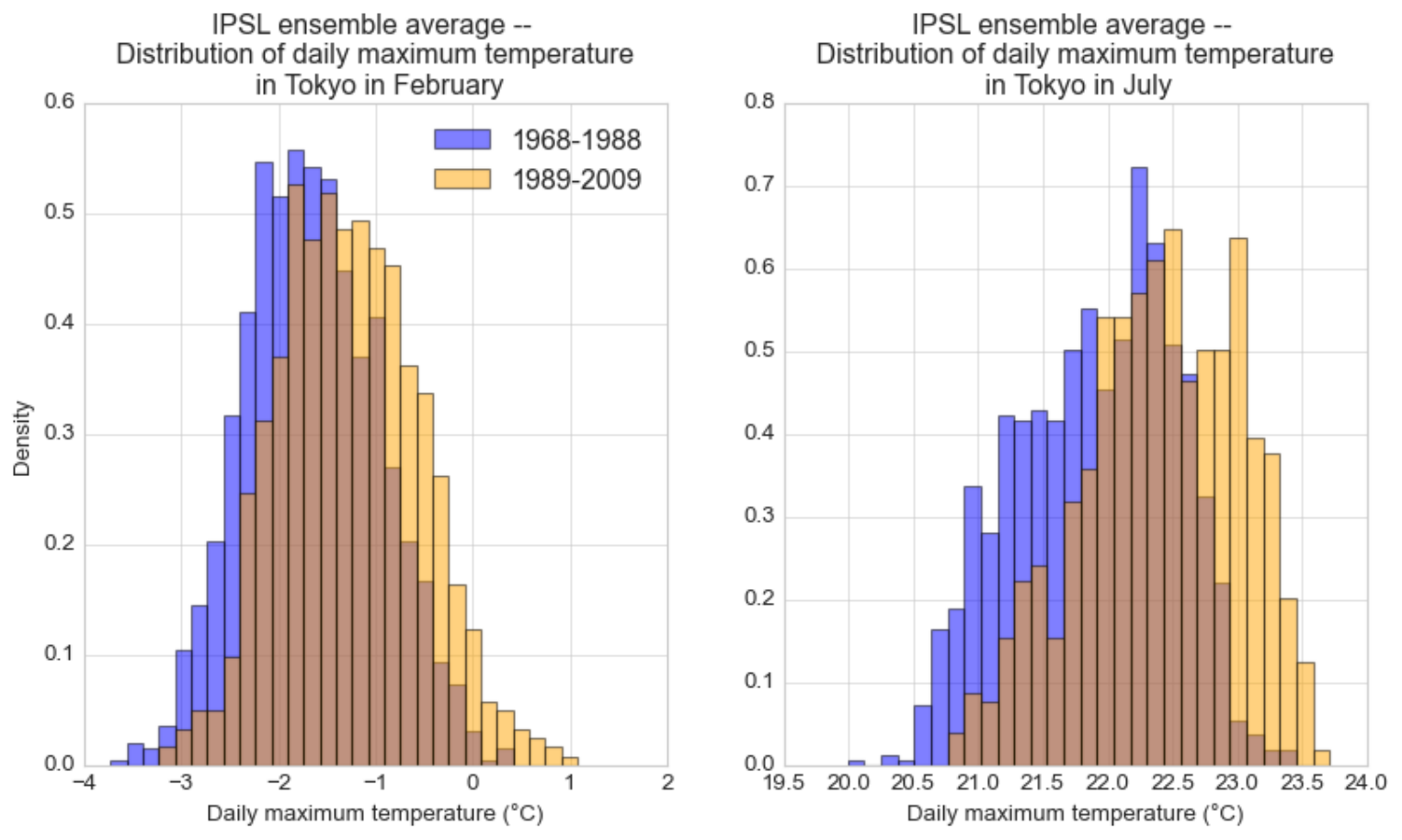}
        \caption[]{The distribution of modelled (IPSL) ensemble average of maximum daily temperatures in Tokyo, Japan had changed between 1968-1988 (blue histogram) and 1989-2009 (orange histogram) in February (left) and July (right). Qualitatively, the change is different than that in the observed record (see Figure \ref{tokyo_obs_feb_july})}
        \label{ipsl_ensemble_avg_daily_max_temp_feb_july_tokyo}
    \end{minipage}\hfill
    \begin{minipage}{0.48\textwidth}
        \centering
        \includegraphics[width=\linewidth]{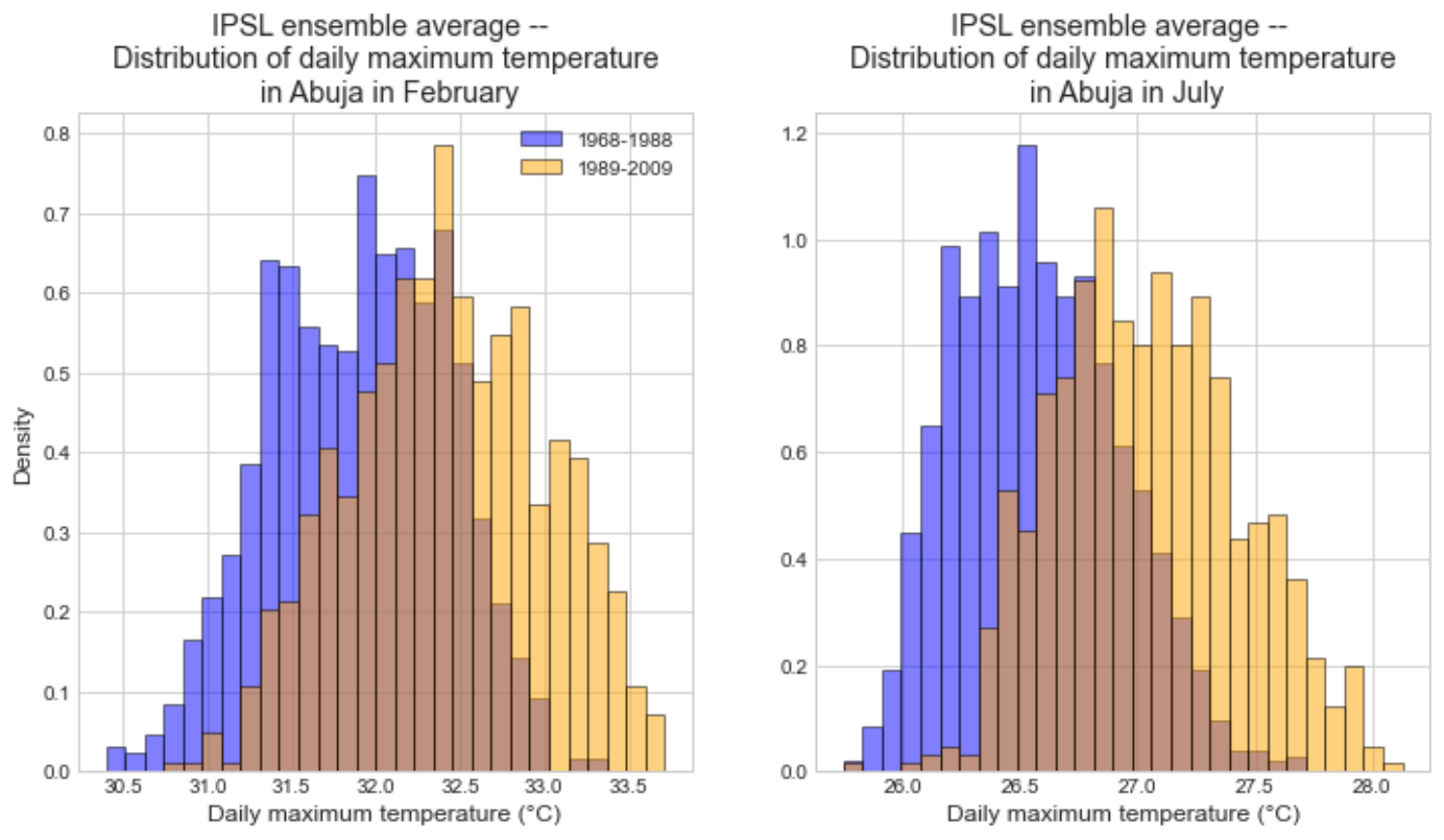}
        \caption[]{The distribution of modelled (IPSL) ensemble average of maximum daily temperatures in Abuja, Nigeria had changed between 1968-1988 (blue histogram) and 1989-2009 (orange histogram) in February (left) and July (right). Qualitatively, the change is different than that in the observed record (see Figure \ref{abuja_feb_daily_temp})}
\label{abuja_ensemble_avg_temp_shift_feb_july}
    \end{minipage}
\end{figure}
\vspace{-10pt} 

\begin{figure}[ht]
    \centering
    \begin{minipage}{0.48\textwidth}
        \centering
        \includegraphics[width=\linewidth]{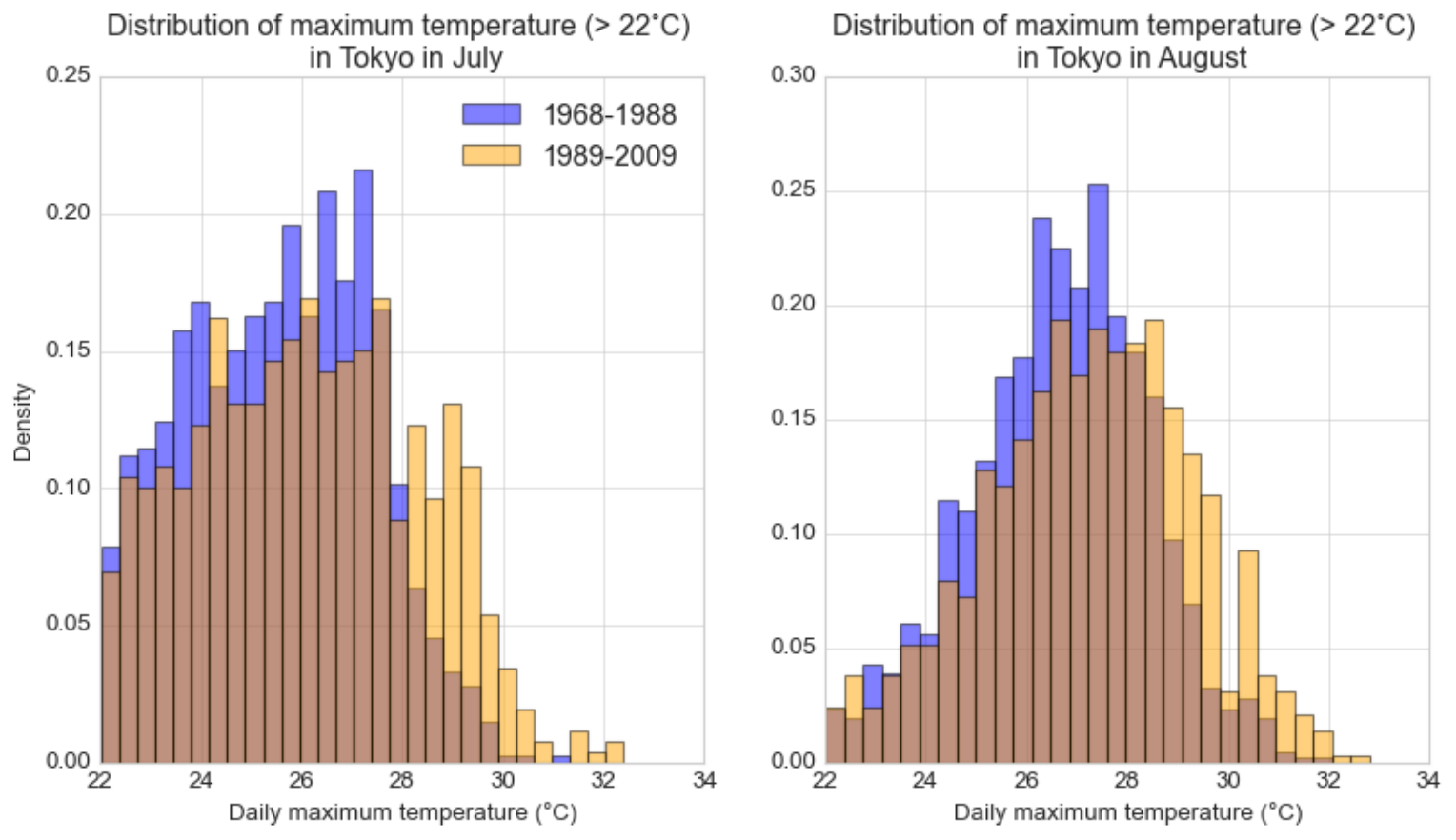}
        \caption[]{The distribution of observed (ERA5) daily maximum temperature above $22 ^\circ C$ in Tokyo had shifted between 1968-1988 (blue histogram) and 1989-2009 (orange histogram) in July (left) and August (right)}
        \label{tokyo_above_22_july_august}
    \end{minipage}\hfill
    \begin{minipage}{0.48\textwidth}
        \centering
        \includegraphics[width=\linewidth]{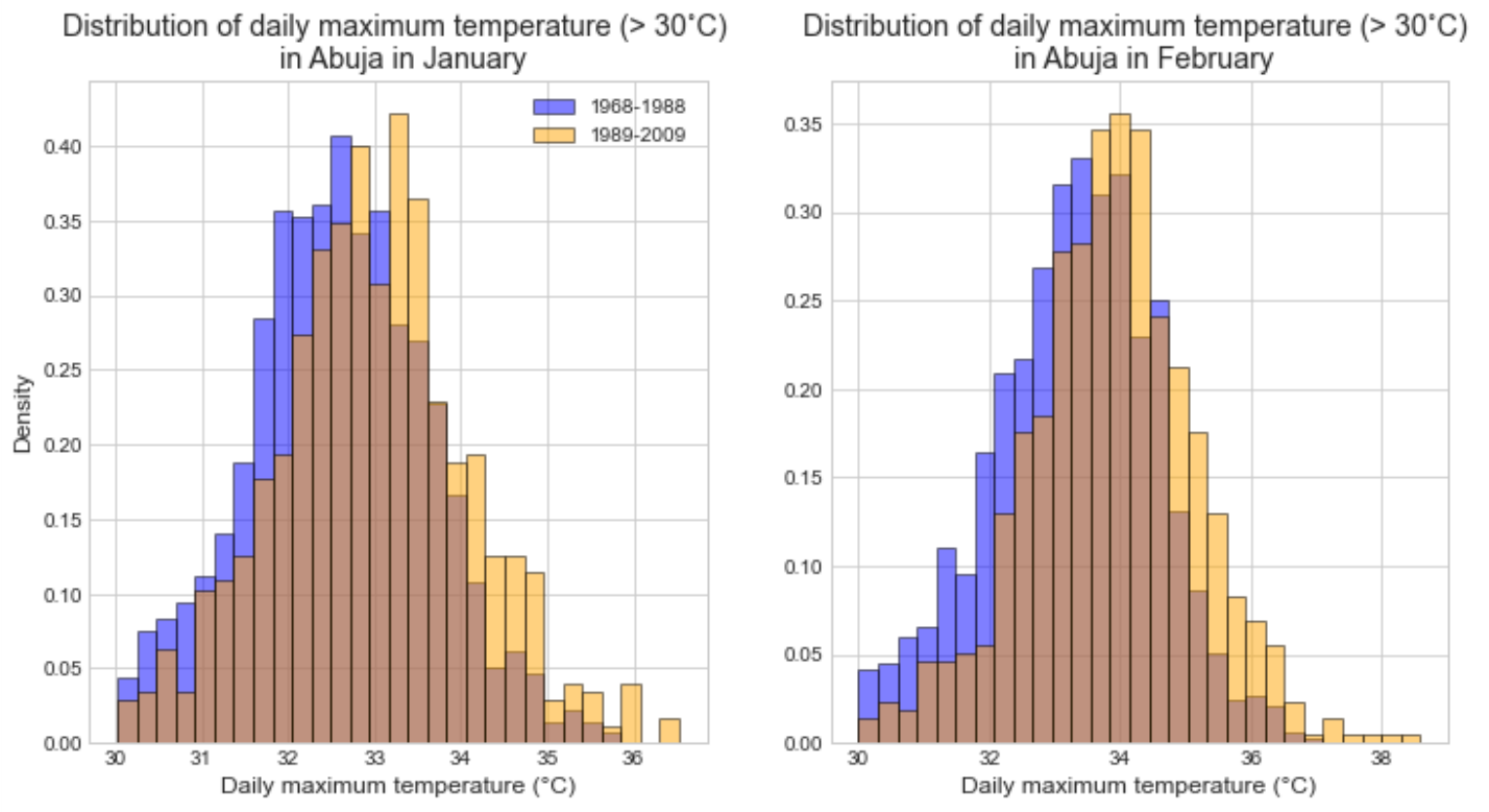}
        \caption[]{The distribution of observed (ERA5) daily maximum temperature above $30 ^\circ C$ in Abuja had shifted between 1968-1988 (blue histogram) and 1989-2009 (orange histogram) in January (left) and February (right)}
        \label{jan_feb_abuja_obs}
    \end{minipage}
\end{figure}

\begin{figure}[H]
    \centering
    \begin{minipage}{0.48\textwidth}
        \centering
        \includegraphics[width=\linewidth]{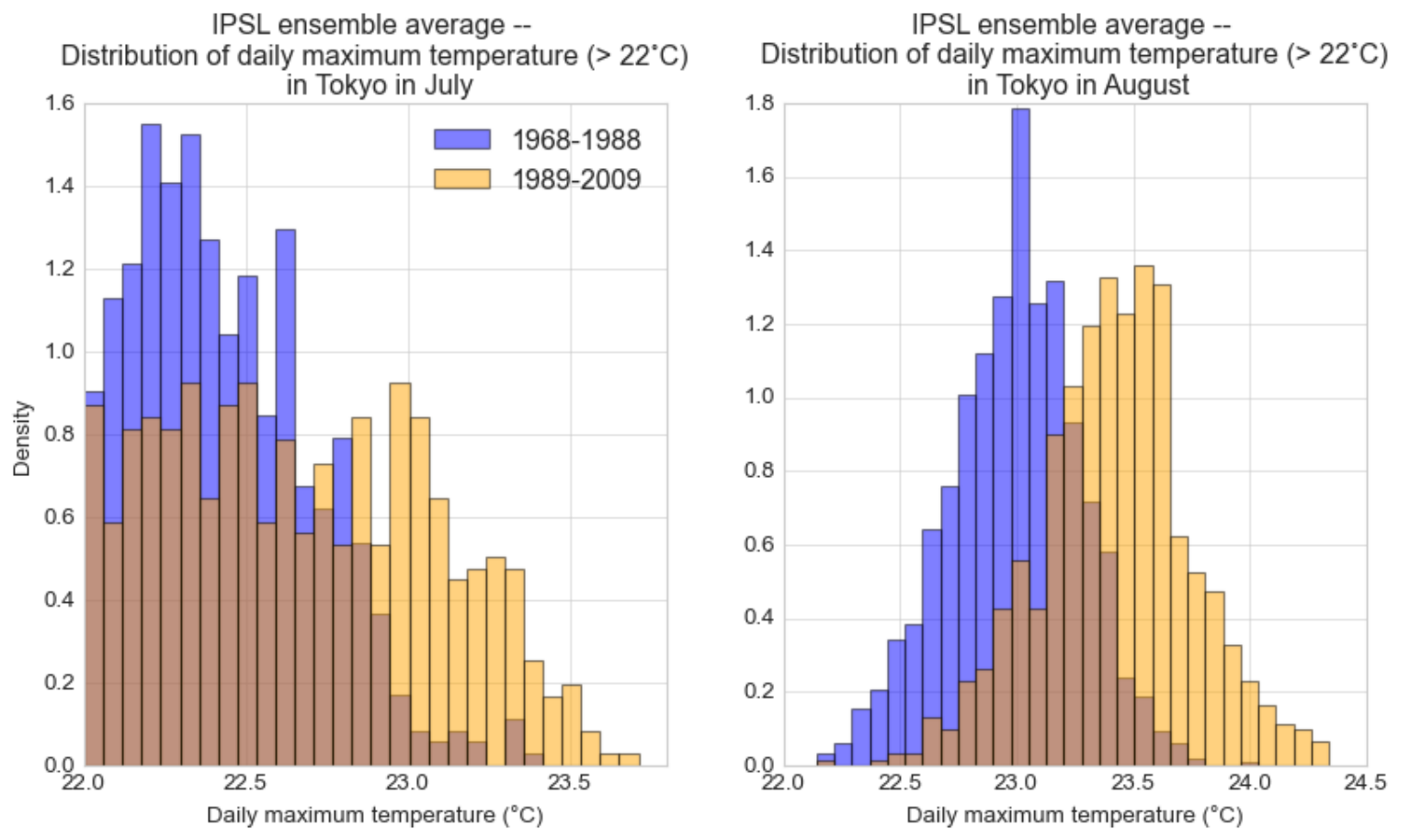}
        \caption[]{The distribution of modelled (IPSL) daily maximum temperature above $22 ^\circ C$ in Tokyo had shifted between 1968-1988 (blue histogram) and 1989-2009 (orange histogram) in January (left) and February (right). Qualitatively, the change is different than that in the observed record (see Figure \ref{tokyo_above_22_july_august})}
        \label{dist_shift_tokyo}
    \end{minipage}\hfill
    \begin{minipage}{0.48\textwidth}
        \centering
        \includegraphics[width=\linewidth]{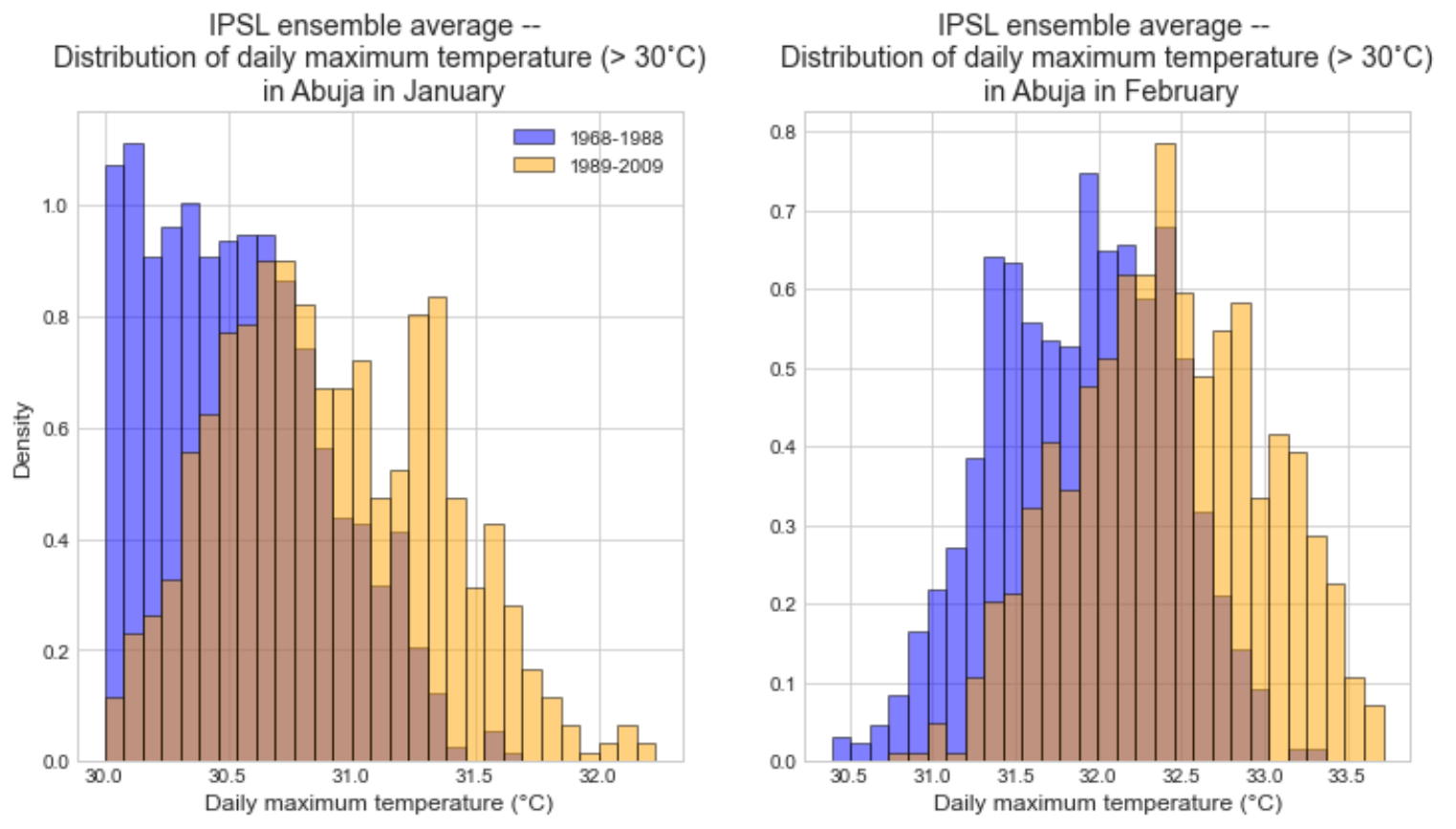}
        \caption[]{The distribution of modelled (IPSL) daily maximum temperature above $30 ^\circ C$ in Abuja had shifted between 1968-1988 (blue histogram) and 1989-2009 (orange histogram) in January (left) and February (right). Qualitatively, the change is different than that in the observed record (see Figure \ref{jan_feb_abuja_obs})}
        \label{ipsl_abuja_30_above_jan_feb}
    \end{minipage}
\end{figure}

\section{Motivation for the Taylorformer temporal BC} \label{motivation}
Here, we present the motivation to reformulate BC as a time-indexed regression task. Further, we discuss why we thought it could address the time asynchronicity between climate models and observations and how it can resolve local-scale variability.     

First we show that some standard BC methods, such as mean shift, can already be written as a time-indexed regression model. However, since their outputs cannot be taken as literal values at a specific time-point, we progress to more flexible models that can.

\subsection{A univariate BC as a time-indexed regression model}
In our method, we use a regression model to perform the BC. However, other BC methods can also be performed by regression. In the following, we examine the mean shift \cite{Xu} (used, for example, by Mora et al. \cite{Mora} for heatwave mortality impacts). 

In a mean shift, observations and climate model outputs are collected over a reference period, and the difference between their means is calculated. The difference is then added to future climate projections (hereafter, we call this execution the 'procedural approach').

Alternatively, let $O_i$ be the random variable representing the temperature readouts at time-point $t_i$ for observations and $g_i$ the corresponding climate model value. Now, the following probability model can be written: 
\begin{equation}
\label{eq::1}
   O_i = g_i + c + E_i  
\end{equation}
\noindent where $E_{i} \sim \operatorname{Normal}\left(0, \sigma^2\right)$ and $E_{i}$ is independent of $E_{j}$ for $i \neq j$ and $c$ is a learned parameter. To find $c$, we can use standard linear regression packages. Standard linear regression uses maximum likelihood as its objective to learn the parameters. We show in Appendix \ref{appendix_LR} the equivalence of the procedural approach and the probability model. As a trivial extension, the mean and variance shift probability model is: 
\begin{equation}
O_i = b \times g_i + c + E_i 
\end{equation}
\noindent where both $b, c$ are learned parameters using standard linear regression (again using the maximum likelihood objective).

This setup of the two equations above is unusual in the BC literature since the index $i$ is the same on both sides of the equation. Generally, BC practitioners avoid such a formulation since we know that climate models and observations are not in temporal synchrony (\cite{maraun_widmann_2018}, p.137). Nevertheless, we can write it in this way and arrive at the same results as the 'procedural approach.' 

This view shows that we can use regression models for BC (and, in a sense, we already do implicitly). For the BC practitioner, it may be of interest that this view also provides a way to treat mean shift and mean-variance shift as stochastic BC methods (here, the stochasticity comes from what is known in statistics and ML literature as aleatoric uncertainty \cite{Hllermeier2019AleatoricAE}). 

After performing the regression, we will get, for each random variable, $O_i$, a distribution with a mean value equaling $g_i + c$. In this regression model, the mean value of $O_i$ at time $t_i$ should not be taken to be the literal value at that exact time-point due to temporal asynchrony. In other words, if tested using MSE on an unseen hold-out data, the value of $\dfrac{1}{M}\sum_{i=1}^{M} \left(o_i - \left[g_i + c\right]\right)^{2}$ will be large, i.e., the model performs poorly.  

The fact that the distribution obtained by $O_i$ at time $t_i$ should not be taken to be the literal values at that exact time-point leads us to search for more flexible regression models where it may be. 

Even if the two time-series were synchronised, it is clear that we cannot use it to correct time-dependent statistics since the random variables $O_i$ and $O_j$ are independent of each other for any $i \neq j$. This is equivalent to throwing away all temporal correlation information and relying solely on the temporal correlation of the climate model. 

\subsection{Outputs with a specific time index can be meaningful}
In weather forecasting, when we say \say{output with a specific time-index is meaningful}, we mean that the estimated value $\hat{o}_i$ of the random variable ${O}_i$ is very close (by MSE) to the observed value $o_i$ and its confidence interval is very narrow. To be 'meaningful,' however, in a time-series context on a longer time horizon, we do not need to have a narrow confidence band or a small MSE value. On a climatic time horizon, we cannot precisely specify the value obtained in Tokyo, Japan, on January 1st, 2050. However, we can still look at the distribution at that specific time point. One added benefit to this approach is that we can account for temporal correlations, i.e., what happened on day $t$ can affect the possible values at day $t+1$.     

\begin{figure}[H]
    \begin{minipage}{0.48\textwidth}
        \centering
        \includegraphics[width=\linewidth]{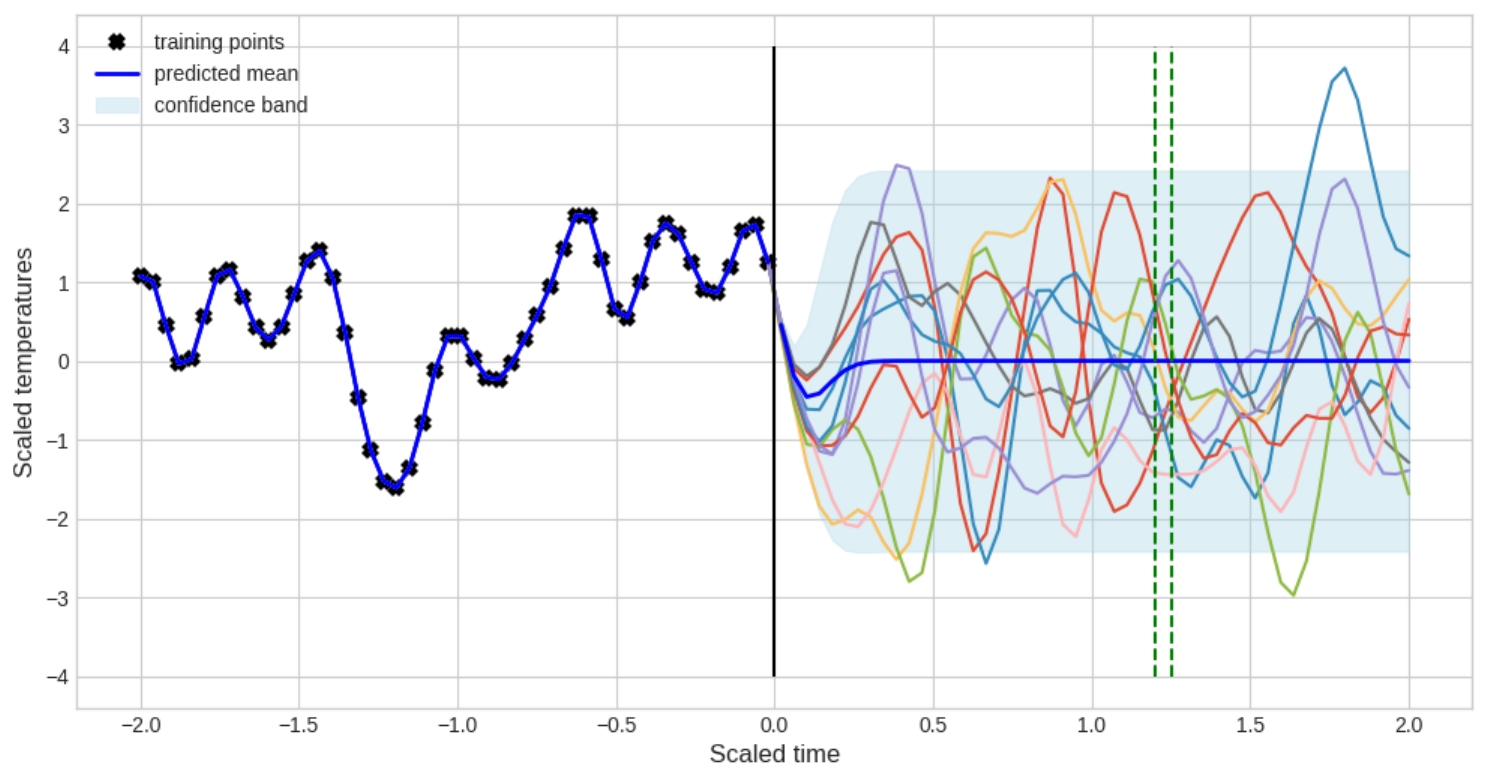} 
        \caption{The task involves predicting future values using training points ('+') located left of the vertical black line. The predicted mean (solid blue line) aligns closely with these points and quickly reverts to zero beyond this boundary. The light blue shading indicates the $95\%$ confidence interval. Sample trajectories from the predicted probability model are shown as colored lines to the right of the black line. These demonstrate that target variables distant from the training data maintain the same marginal distribution, mean, and variance, as evidenced by the vertical slice marked by green dashed lines. This time-series view highlights correlations among points, offering insights not visible from a simple histogram of training points.}
        \label{GP_time_is_meaningful}
    \end{minipage}\hspace{0.04\textwidth}
        \begin{minipage}{0.48\textwidth}
        \centering
        \includegraphics[width=\linewidth]{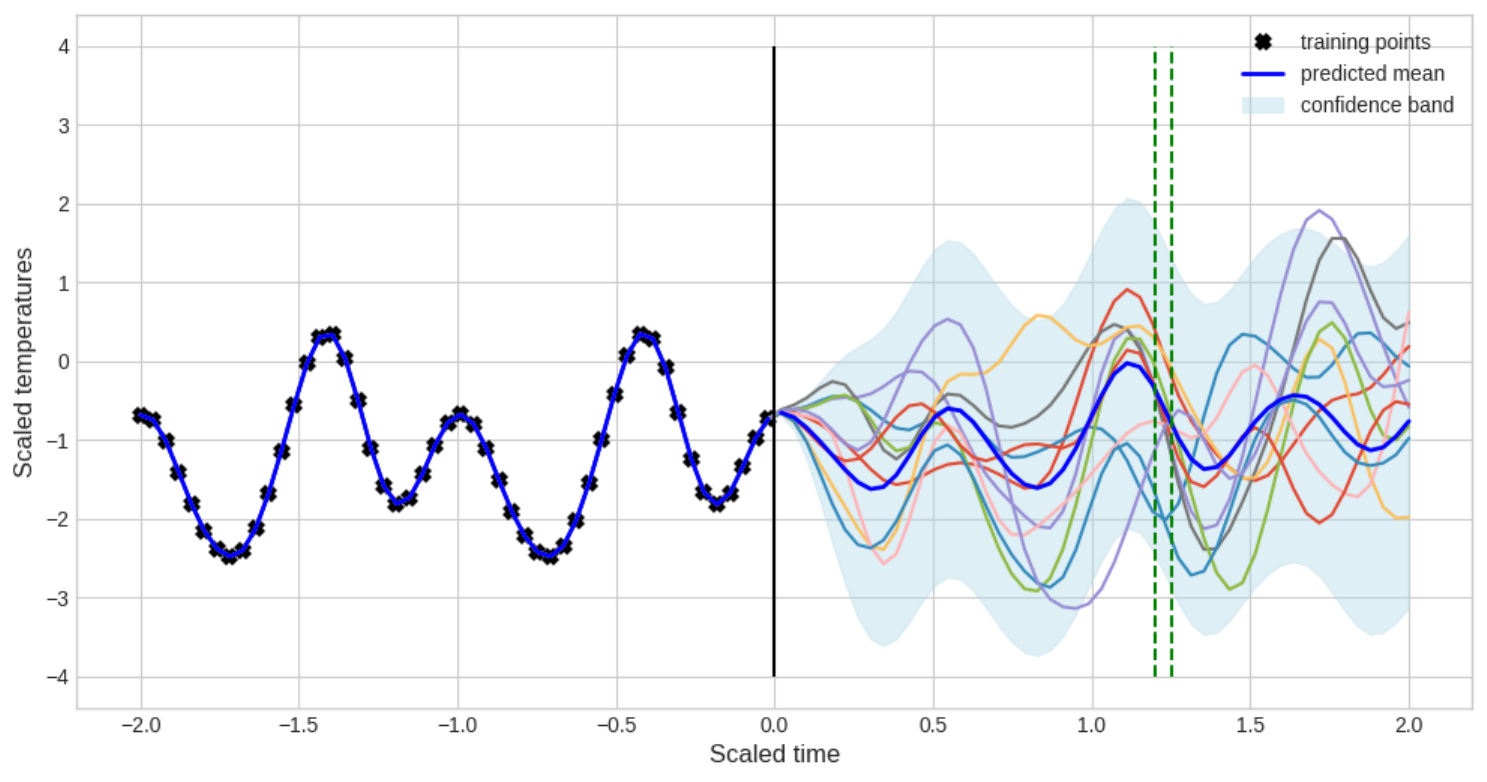} 
        \caption{Predictions are made using training points ('+') left of the vertical black line. The predicted mean (solid blue line) aligns accurately with these points and, upon extrapolation to the right, adapts to typical 'seasonality' behaviors instead of becoming non-informative (refer to Figure \ref{GP_time_is_meaningful}). The light blue shading indicates the $95\%$ confidence interval for each x-axis value. Various sample trajectories from the probability model are shown as colored lines, while the dashed green lines illustrate the marginal distribution at corresponding x-axis values}
        \label{GP_seasonal}
    \end{minipage}
\end{figure}

The fact that some probability models can be useful for long-time horizons, even when looking at a specific time index, is not specific to BC. As an illustrative example, we will discuss a general prediction task for time series using a Gaussian Process (GP) regression model. The GP is specified by what is known as a Radial Basis Function (RBF) kernel. In Figure \ref{GP_time_is_meaningful}, we show that when extrapolating further from the training data points ('+'), our mean prediction (solid blue line) converges to zero. Let us look at the marginal distribution of the target variable $Y_{80}$ with $t_{80} = 1.2$ (roughly, data between the dashed green lines). We will get $Y_{80} \sim \operatorname{Normal}(0, 1.52)$, and $Y_{90} \sim \operatorname{Normal}(0, 1.52)$ as well. There are two takeaways from this model: (1) random variables further away from the data will revert to the typical behaviour (mean zero), and their confidence interval will widen in comparison to regions close to the training data. This is seen clearly in the region to the left of the vertical black line (the X-axis value is zero) and to its right. However, they still provide useful information, namely the typical behaviour, when viewed with a specific time-index. And (2) while the marginal distribution is the same correlation information shown by the wiggly lines to the right of the vertical black line.

As a second example, consider another GP shown in Figure \ref{GP_seasonal}: the random variables' mean (solid blue line) far away from the training data (`+') again reverts to the typical behaviour. However, in this case, the typical behaviour includes the 'seasonality' seen in the past. (See Appendix \ref{GP_appendix} for the exact details of the two GP models above.)

These are two pictures that serve as a motivation for the temporal BC task. Similarly to the pictures of the GP, we would expect the marginal distribution of random variables far away from the training data to describe a typical behaviour such as seasonal, yearly or decadal means. But, an additional benefit would be learning the correlation structure between variables. To learn the correlation structure, we will need a regression model that allows for dependence between points (as opposed to a mean shift or an EQM model).   

\subsection{Time indexed BC can correct systematic time asynchronicities}
\begin{figure}[H]
        \centering
        \includegraphics[width=\linewidth]{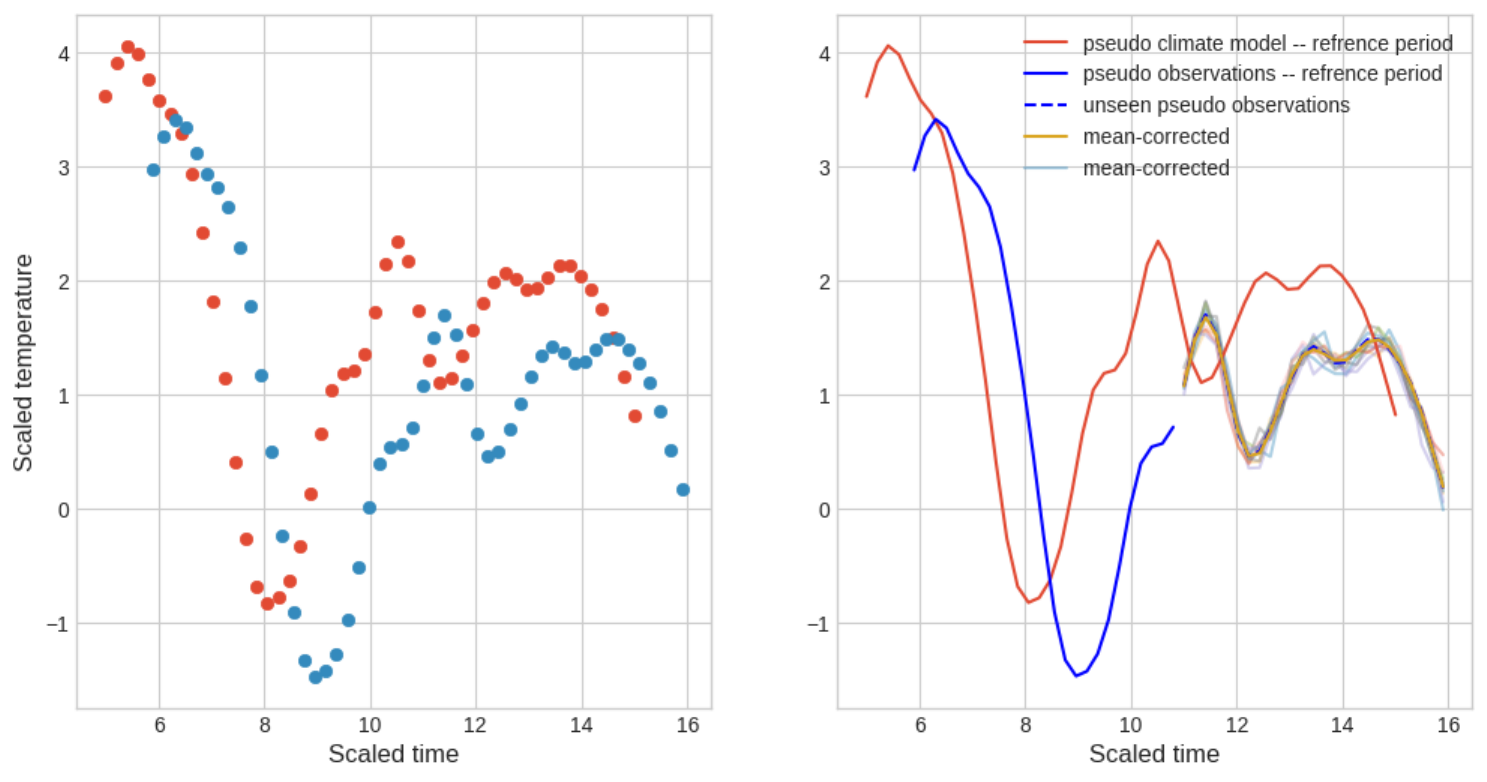} 
        \caption{This figure uses a toy example of a Gaussian Process-based BC regression (details in Appendix \ref{GP_appendix}) to illustrate correction of systematic time asynchronicities between 'climate model' and observations. Left panel: The pseudo observations (blue points) are both vertically and horizontally shifted relative to the pseudo climate model (red points). Right panel: The GP model's mean (golden solid line) successfully captures the vertical and horizontal shifts, as shown by the alignment of the pseudo observations (solid blue line) with the pseudo climate model's full time-series (solid red line). Traditional quantile mapping approaches with a fixed reference period would fail to recognize that the processes are identical, barring a mean shift}
        \label{time_misplacement_fig}
\end{figure}

First, let us look at a simple toy example of time asynchronicity. In the left panel of Figure \ref{time_misplacement_fig}, we show two identical time series in which we added a mean bias to the pseudo observations (blue points) and then horizontally shifted the whole time-series to the right. Here, The horizontal shift is the 'time asynchronicity'. The red points represent the pseudo-climate model data. If we were to use 1D histograms over the same reference period, as is typical in univariate BC, we would not be able to infer that the two time-series are identical. Differently, a time-indexed probability model like a GP can learn the time-shift from data, as shown in the right panel of Figure \ref{time_misplacement_fig}. Note that the input to the GP includes both (past and future) pseudo-climate model and past observations to make the correction possible.  

As a second example, imagine the heatwave duration is consistently two times longer in the climate model than in the observational record. We can then manually construct a probability model. Our probability model copies the climate model time-series; if the time-series passes above $30 ^\circ C$ for multiple days in a row ('heatwave'), cut the 'heatwave' time-series in half and add a bias $a$ and some noise, $E_i$, to get the corresponding estimated observation; otherwise, assign 0. 

The mathematical notation for the associated probability model is not important here, but the main takeaway is that the value at time point 
$t_{i}$, $o_i$, is some known deterministic function $f$ of the (past, future) values from the climate model. Since we added bias $a$ and a noise term we can write the model mathematically as 
\begin{equation}
\label{eq:myEquation}
O_k|\textit{inp}_{k} \sim \operatorname{Normal}\left(f\left(\textit{inp}_{k}\right) + a, \sigma^{2}\right)
\end{equation}
\newline
\noindent where $\textit{inp}_{k} = \left(g_1, \dots g_M, k\right)$. That is, again, an easy probability model with a learnable parameter $a$ which we can learn using standard linear regression libraries.

If in addition to the heatwave duration setup, we have a consistent mean shift of one degree ($^\circ$C) between the climate model and observations, we can learn this shift by adding to the input, $\textit{inp}_{k}$, past observations, $o_1 \dots, o_n$, where $n <K$. From this example, we see that the values from the climate model in the (past, future) and past observations will be useful in the probability model we are constructing.  

\subsection{Time indexed BC coupled with downscaling can resolve local scale variability}
If the resolution of the climate model is coarser than that of the observed dataset, one grid point in the climate model is the average behaviour of multiple grid points in the observed data. If we use EQM or mean shift, for example, high values in the projections of a GCM will map to high estimated observed values, regardless of the specific behaviour at the corresponding grid-point in the observed data. For quantile-mapping approaches, this is also known as the inflation issue \cite{Maraun2013BiasCQ}.(See more details in Maraun and Widmann, 2018, p.189-192 \cite{maraun_widmann_2018}.)

To have a model that can differentiate between the local scale grid points, we can simply add the time-series of the observed values as input. This is another reason for adding past observations to our regression model.       

\subsection{Tying it all together}
In the context of BC, we can perform a regression model. Due to a lack of temporal synchrony, a regression model (like mean shift) with one-to-one correspondence would perform poorly on a hold-out set, and its values should not be taken literally. However, a regression model with a specific time index that does not assume independence between target variables, like a GP (or other classic AR model), can a) describe the typical distributional behaviour of points far away from the training data without throwing away temporal information, and b) since it receives as input many values from different time-points (past, future) climate model and past observations can find patterns that are not a one-to-one correspondence and potentially fix time asynchronicities (such as in Figure \ref{time_misplacement_fig}). 
Furthermore, suppose the climate model is coarser in resolution than the observed data (one climate model grid point to many observed grid points). In that case, the regression model can potentially differentiate between the grid points and adapt to local variability if its input includes past observations.    

We can write the GP version of the temporal BC probability model as 
\begin{equation}
\label{prob_model_app}
    O_k|\textit{inp}_{k} \sim \operatorname{Normal}\left(f\left(\textit{inp}_{k}\right), g\left(\textit{inp}_{k}\right)\right)
\end{equation}

here $O_k$ is a random variable representing the daily maximum temperatures at time $t_k$ for observations, where we have $ t_i < t_j$ if $i < j$ and $M$ being the size of the sequence. Furthermore, we denote with $t'$ the time-points of the climate model, that are potentially different than the observed time-points.       
Additionally, here $\textit{inp}_{k} = \left(\left(g_1, t'_1\right), \dots \left(g_M, t'_M\right), \left(o_1, t_1\right), \dots \left(o_{k-1}, t_{k-1}\right)\right)$, $f$ and $g$ are functions of the data defined by the GP.  

If the systematic patterns of time asynchronous were clear to the human eye or by standard analysis, we could construct a GP model with the appropriate equations. However, since for the BC task, we do not know how to specify these interactions, we turn to a more flexible neural network model that can learn these interactions from the data. 

\section{Linear regression mean shift derivation}\label{appendix_LR}
Estimation of the parameters in linear regression is done using maximum likelihood as the objective. Using maximum likelihood and denoting $\textbf{g}$ to refer to the set of all climate model outputs, we have 
\begin{equation}
    \begin{aligned}
        & \max_{\theta} P\left(O_1, \dots, O_N \middle| \mathbf{g}\right) 
        \overset{\text{independence}}{=} \prod_{i=1}^{N} P(O_i|g_i) \\
        &\overset{\text{normal r.v.}}{=} \prod_{i=1}^{N} \dfrac{1}{\sqrt{2\pi\sigma^{2}}}\exp\left(- \dfrac{(o_i - [g_i - c])^{2}}{2\sigma^{2}}\right) \\
        &\xrightarrow{\text{log}} \sum_{i=1}^{N}\log(1) - \log\sqrt{2\pi\sigma^{2}} - \dfrac{(o_i - [g_i - c])^{2}}{2\sigma^{2}} \\
        & \dfrac{\partial}{\partial c} \xrightarrow{\text{derivative} = 0} - \sum_{i=1}^{N}\left((o_i - [g_i - c])\right) = 0 \\
        &= \dfrac{1}{N}\sum_{i=1}^{N} o_i - \dfrac{1}{N}\sum_{i=1}^{N} g_i
    \end{aligned}
\end{equation}

\section{From Transformer decoder to Taylorformer} \label{Taylorformer_appendix}

\subsection{Transformer decoder for language modelling}
One component in recent successes of large language models, like GPT \cite{Brown2020LanguageMA}, is their Transformer decoder architecture. We will now give an abstract description of the architecture --- the specifics can be found in \cite{Liu2018GeneratingWB}, but the abstract view, we believe, makes it easier to explain the salient parts of the model and makes it clear which parts can be replaced with ease.     

For language modelling our dataset is comprised of $N$ sequences, each comprised of (word, position) pairs ($W_i$, $P_i$) for $i \in 1 \dots K_s$. $K_s$ is the length of sequence $s$. Both $W_i$ and $P_i$ are scalars. We will take $D$ to denote the total number of unique words in our dataset.  

The goal of language modelling is to classify the next word in a sentence based on previous words. 

\subsubsection{Feature Engineering}
In the raw data, $P_i$ is a scalar. One of the ingenious of the Transformer decoder is the extraction of features from $P_i$; these are prepared before training and are inspired by Fourier analysis. The formula is:
\begin{align}
    & P'\left( i, 2l\right) = \sin\left(\dfrac{P_i}{\left(10000\right)^{2l/d}}\right)\\
    & P'\left(i, 2l+1\right) = \cos\left(\dfrac{P_i}{\left(10000\right)^{2l/d}}\right)
\end{align}
\noindent $l$ is the dimension and after the features are extracted $P'_i \in \mathbf{R}^d$.

\subsubsection{One attention layer}
As a first step, we map each word $W_i$ to a vector of learnable weights $W'_i \in \mathbf{R}^h$. $W'_i = W'_j$ if and only if $W_i = W_j$. Secondly, we concatenate $W'_i$ and $P'_i$ and pass them through three separate non-linear functions (could be multiple layers of a neural network) $f_\theta$, $g_\theta$ and $h_\theta$, let's call the resulting vectors $Q_i, K_i, V_i$ all of which are in $\mathbf{R}^{d'}$. 

\paragraph{Attention weights} are then simply constructed as 
$A_{i, j}$ = $\operatorname{Softmax_{j\in {1, \dots, i}}}\left(\sum_{\ell=1}^{d'}Q_{i,\ell}K_{\ell, j}\right)$

\paragraph{Prediction} for the probabilities for all classes (all possible words in the dataset) for target variable $W_i$ after one layer can then be made by $\pi_i\left(\theta\right) = \operatorname{Softmax_{d\in {1, \dots, D}}}B_\theta\left(\sum_{j=1}^{i-1}A_{i-1, j}\times V_{j}\right)$. $B_\theta: \mathbf{R}^{d'} \rightarrow \mathbf{R}^D$ is a non-linear function with dimension output equal to the number of unique words in the dataset $D$, and $\pi_i\left(\theta\right) \in \mathbf{R}^{D}$.
Here, we assumed the target word is distributed according to a $\operatorname{Categorical}$ random variable with $D$ categories. We can write it as $W_i \sim \operatorname{Categorical}\left(\pi_{i, 1}\left(\theta\right), \dots \pi_{i, D}\left(\theta\right)\right)$      

\subsubsection{Multiple attention layers}
Each attention layer after the first one assumes the same structure: take the output (before the last $\operatorname{Softmax}$ operator) from the previous layer to be the new values for vectors $Q_i, K_i, V_i$ and repeat the calculation of attention weights and prediction (excluding the last $\operatorname{Softmax}$). In the last layer add the $\operatorname{Softmax}$ to get a probability for a word at position $i$. 

\subsubsection{Multiple attention heads}
It is common to split the procedure within each attention layer into multiple heads (or branches). This works by taking the vectors $Q_i, K_i, V_i$ and splitting them into $H$ heads such that $V_i$ is equal to the concatenation of $V_i^h$ for $h \in 1, \dots H$ and $V_i^h \in \mathbf{R}^{d' / H}$ and a similar split is performed for $K_i$ and $Q_i$. Then we can follow the same procedure as before for attention weights for each head $h$ to get $A_{i, j}^{h}$. After we calculate $\sum_{j=1}^{i-1}A_{i-1, j}^h\times V_{j}^h$, we can concatenate all the heads and continue as before.

\subsubsection{Training objective}
The training objective will be to maximise the log-likelihood of the data. For one sentence with words $W_1, \dots W_{K_s}$ this means, 
\[
\operatorname{maximise}_{\theta} \operatorname{Log} P\left(W_{1}, \dots, W_{K_s}, \theta\right) \underbrace{=}_{\text{chain rule}}\operatorname{maximise}_{\theta} \operatorname{Log} \left( P(W_{K_s}|W_{1:{K_s-1}}, \theta) \cdots P(W_{2}|W_{1}, \theta) P(W_{1}| \theta) \right)
\]

\noindent, where we have used the chain rule to decompose the joint probability and $\theta$, refers to the parameters we want to learn via gradient descent using our model. Now, we just need to replace each term in the equation with the categorical distribution. If the true word, $W_i$ at position $i$ is $w_i \in \mathbf{N}$, we will denote its predicted probability as $\pi_{i, w_i}$, then we can write the log-likelihood maximisation objective as
\begin{equation}
\operatorname{maximise}_{\theta} \operatorname{Log}
\pi_{1, w_1}(\theta) \times \dots \times \pi_{K_s, w_{K_s}}(\theta)
\end{equation}

\subsubsection{Inference}
At the inference stage, we can simply fix $W_{1} = 0$ where the $0$-th index indicates the first word being the token \textless sos\textgreater (start-of-sentence), we then do a forward pass of our model to get $\pi_2$, we then sample $\left(W_{2}|W_{1} = 0\right) \sim \operatorname{Categorical}\left(\pi_{2, 0}\left(\theta\right), \dots, \pi_{2, D}\left(\theta\right)\right)$. Next, we make another forward pass now with the fixed value of $W_1$ and the sampled value of $\left(W_{2}|W_{1} = w_1\right)$ to get $\pi_3\left(\theta\right)$. We continue this procedure until the output of the sampling gives us the token \textless eos\textgreater (end-of-sentence).   

\subsubsection{Language sequences versus Continuous Processes}
Language sequences follow a discrete regular grid --- at each position $p$ we are interested in the word at position $p+1$ as opposed to predicting $k$ steps ahead, i.e., what will be the word at position $p+k$ given the words up to position $p$. For this reason in the equations above, when predicting the next word, we don't require information about the position to be predicted next. This situation may fit some continuous processes, but there are other situations for continuous processes, like interpolation, long-term prediction and missing values, that would not work well with the same setup. We first introduce the Transformer decoder for regular grid continuous processes and then extend it to the more general case.      

\subsection{Transformer decoder for regular-grid continuous processes}
A simple adaptation is needed for translating the Transformer decoder from a language model to a continuous processes model. Briefly, we just need to adapt the feature engineering, remove the $\operatorname{Softmax}$ operator from the prediction stage and choose a different distribution for our targets. We do not need to change anything about the 'Multiple attention layers' and 'Multiple heads' sections.  

\subsubsection{Feature Engineering}
In the raw data now we have time-point, $t_i \in \mathbf{R}$, instead of a position, $P_i \in \mathbf{N}$. The adapted formula we suggest is:
\begin{align}
    & P'\left(t_i, 2l\right) = \sin\left(\dfrac{t_i/\delta_t}{\left(t_{\text{max}}/\delta_t\right)^{2 t_i/d}}\right)\\
    & P'\left(t_i, 2i+1\right) = \cos\left(\dfrac{t_i/\delta_t}{\left(t_{\text{max}}/\delta_t\right)^{2 t_i/d}}\right)
\end{align}
\noindent So after the features are extracted $P'_i \in \mathbf{R}^d$. Where $t_{\text{max}}, \delta_t$ are hyperparameters. 

\subsubsection{One attention layer}
In the continuous case, we have real-valued random variables $O_i$  with their observed values being $o_i$ at time-point $t_i$. In language modelling we mapped the index of word $W_i$ to a vector of learnable weights. Here, we can just pass each value $o_i$ such that $O'_i \in \mathbf{R}^h$, where $O'_i = A_\theta(o_i)$ where $A_\theta$ denotes non-linear neural network layers. Then, we proceed as before by concatenating $O'_i$ and $P'_i$ and pass them through three separate non-linear functions (could be multiple layers of a neural network) $f_\theta$, $g_\theta$ and $h_\theta$, let's call the resulting vectors $Q_i, K_i, V_i$ all of which are in $\mathbf{R}^{d'}$.

\paragraph{Attention weights} are calculated in exactly the same manner as before, i.e., $A_{i, j}$ = $\operatorname{Softmax_{j\in {1, \dots, i}}}\left(\sum_{\ell=1}^{d'}Q_{i,\ell}K_{\ell, j}\right)$

\paragraph{Prediction} for the continuous case differs only in the last layer; instead of a last layer of $\operatorname{Softmax}$ to get class probabilities (used in the language model), we can use for the last layer a linear layer $B_\theta \mathbf{R}^{d'} \rightarrow \mathbf{R}^2$, such that \mbox{$\mu_i\left(\theta\right), \sigma_i\left(\theta\right) = B_\theta\left(\sum_{j=1}^{i-1}A_{i-1, j}\times V_{j}\right)$}. 
Here, we assumed the target random variable is distributed
according to a $\operatorname{Normal}$ random variable with two hyperparameters $\mu, \sigma$. We can write it as $O_i \sim  \operatorname{Normal}\left(\mu_i\left(\theta\right), \sigma_i\left(\theta\right)\right)$.

\subsubsection{Training objective}
The training objective will be to maximise the log-likelihood of the data. For one sequence with observations $O_1, \dots O_{K_s}$ this means, 

\[
\operatorname{maximise}_{\theta} \operatorname{Log} P\left(O_{1}, \dots, O_{K_s}, \theta\right) \underbrace{=}_{\text{chain rule}}\operatorname{maximise}_{\theta} \operatorname{Log} \left( p(O_{K_s}|O_{1:{K_s-1}}, \theta) \cdots p(O_{2}|O_{1}, \theta) p(O_{1}| \theta) \right)
\]

\noindent, where we have used the chain rule to decompose the joint probability and $\theta$, refers to the parameters we want to learn via gradient descent using our model. Now we just need to replace each term in the equation by the $\operatorname{Normal}$ distribution. If the true value, $O_i$ at position $i$ is $o_i \in \mathbf{R}$, we will denote its estimated $\operatorname{Normal}$ probability hyperparmaters as $\mu_i (\theta), \sigma_i (\theta)$, then we can write the log-likelihood maximisation objective as
\begin{equation}
\operatorname{maximise}_{\theta} \operatorname{Log}
\Sigma_{i=1}^{K_s} \dfrac{1}{\sqrt{2\pi \sigma_i (\theta)}}\exp{-\dfrac{\left(o_i - \mu_i\left(\theta\right)\right)^2}{2\sigma_i(\theta)^{2}}}
\end{equation}
\noindent And this could be further simplified to give

\begin{equation}
\operatorname{maximise}_{\theta} \sum_{i = 1}^{K_s}\operatorname{Log}\left(\frac{1}{\sqrt{2\pi \sigma_i (\theta)}}\right) -\frac{\colorbox{yellow}{$\left(o_i - \mu_i\left(\theta\right)\right)^2$}}{2\sigma_i(\theta)^{2}} \text{ where } \colorbox{yellow}{\text{yellow box indicates the squared error}}
\end{equation}
\noindent If we choose our model to output just the mean in the prediction layer, the training objective becomes the $\operatorname{MSE}$. Note that we could choose to provide our model with initial 'information' (content) by fixing few of the random variables to their observed values and then maximising the likelihood conditioned on the content -- this is what we do in Bias-Correction.  

\subsubsection{Inference}
At the inference stage, we can simply fix $O_1$ to the last observed value during training (this is just one option out of many), we then do a forward pass of our model to get $\left(\mu_2\left(\theta\right), \sigma_2\left(\theta\right)\right)$, we then sample $\left(O_2|O_1 = o_1\right) \sim \operatorname{Normal}\left(\mu_2\left(\theta\right), \sigma_2\left(\theta\right)\right)$. Next, we make another forward pass now with the fixed value of $O_1$, ${o_1}$, and the sampled value of $\left(O_2|O_1 = o_1\right)$ to get $\left(\mu_3\left(\theta\right),\sigma_3\left(\theta\right)\right)$. We continue this procedure until desired. 

\subsection{Transformer decoder for irregular-grid continuous processes}
In a Transformer decoder for a regular grid, we described the 
inner part of the prediction (hereafter "weighted attention") equation as 
$B_\theta(\sum_{j=1}^{i-1}A_{i-1, j}\times V_{j}$). Notice that this is a prediction for the $i$-th timestamp, but none of the elements involved in the equation has the data for the next timestamp. 

To adapt it to the continuous case, we can a) add different inputs to the $Q_i$ vector and b) use "weighted attention."

\paragraph{Adapted $Q_i$}
The input to $Q_i$ will be the same as before for values at time-points $t_i$ we want to provide the model in advance and a concatenation of $P'_{i'}$ and $0$ for values at other time-points $t_{i'}$ we wish to predict, i.e., we mask the associated value of $O'_{i'}$.  

\paragraph{Adapted "weighted attention"}
The equation of the "weighted attention" for $O_i$ will now be 
$B_\theta(\sum_{j=1}^{i-1}A_{i, j}\times V_{j}$). Notice that we have changed $A_{i-1, j}$, which did not contain information about the next timestamp, to $A_{i, j}$, which contains the information about $t_i$ through multiple non-linear layers leading to the vector $Q_i$. The sum still runs up to $i-1$ since $K_i$ and $V_i$ stayed the same and did contain information about the actual value of $O_i$, which we do not want to leak. Since $Q_i$ contains information about the next desired time-point, we can ask for a long-term prediction in the future or in the past, and we do not need the time-points to be equally spaced. 

\subsection{Taylorformer decoder}
In the Taylorformer, we have three key differences to the continuous Transformer decoder: 1) a new prediction layer, 2) different inputs to the vectors $Q_i, K_i, V_i$, and 3) an additional attention mechanism. 

\paragraph{new prediction layer}
Instead of predicting the hyper-parameters ($ mu_i (theta) $, $ sigma_i (theta) $) for $ O_i$, we predict the parameters ($ mu_i (theta) $ + $ O_n (i) $, $ sigma_i (theta) $), where $ O_n (i) $ is the value obtained at the closest index to $ i$, $ n (i) $, which was already observed. This was found to empirically ease the learning task since, loosely, the network does not need to start "from scratch." 

\paragraph{Inputs to $Q_i, K_i, V_i$ in the first attention layer}
It is useful (empirically) to add the following features: 
\begin{itemize}
\item empirical difference: $\Delta_i = O_i - O_{n(i)} $
\item distance: $\Delta_{X_i}  =  t_i - t_{n(i)}$
\item empirical derivative: $d_i = \dfrac{\Delta_i}{\Delta_{X_i}}$
\item closest value: $O_{n(i)}$
\item closest time-point: $t_{n(i)}$
\end{itemize}

So the inputs to the non-linear layers leading to $K_i, V_i$ will be the concatenation of $\left(P'_i,\Delta_i, \Delta_{X_i}, d_i, O_{n(i)}, P'_{n(i)}\right)$. and the inputs leading to $Q_i$ will be the concatenation of $\left(P'_i, 0, \Delta_{X_i}, 0, O_{n(i)}, P'_{n(i)}\right)$.

These two parts of the Taylorformer give the model a flavour of a Taylor approximation since we predict the current value as the closest value plus some function of the distance and derivative of closest values.  

\paragraph{Additional attention mechanism}
is constructed in the Taylorformer (hereafter multi-head attention X, or 'MHA-X'). multiple non-linear layers will be applied to $P'_i$ leading to the two vectors $Q^x_i$ and $K^x_i$. The vector $V^x_i$ will be the result of applying a few non-linear layers only on $O_i$. The "additional weighted attention" from this branch is calculated as $B^x_\theta(\sum_{j=1}^{i-1}A^x_{i, j}\times V^x_{j}$). Here, $A^x_{i, j}$ is the dot-product of $Q^x_i$ with $K^x_j$. The "additional weighted attention" from the first layer is then used as the updated values for the vectors $Q^x_i, K^x_i$. For $V^x_i$ at each layer, the input stays the same as in the first layer. This attention mechanism has been shown to empirically help learning in the Taylorformer paper. 

\paragraph{Connecting the two attention mechanisms}
In the last layer, before prediction, we concatenate the outputs from the 'classic' and additional attention mechanisms.  

\subsection{Inputs for a BC task using Taylorformer decoder}
Here, the main adaptation needed is to take the two time-series, the observations and climate model outputs, and fit them into the 1-dimensional formulation presented above. 

There is a notational convenience we can use here, which will uncover that what we refer to as the correction task is just a special case of prediction--- once this becomes apparent, it is clear that all the standard ML tools for prediction can be translated into tools for correction. We define two-dimensional coordinates that encode both the time-point and which time-series we are looking at,

\[
\bigl[x_1,\dots,x_N\bigr] = \bigl[(t'_1,2),\dots,(t'_n,2),
(t_1,1),\dots,(t_m,1)\bigr]
\qquad
\text{where }N=m+n
\]
and let the \say{generalized} time-series be
$\left[ y_1, \dots, y_N \right] = \left[ g_1, \dots, g_n, o_1, \dots, o_m \right]$

The task is then to predict $\hat{y}_1,\dots,\hat{y}_\ell$ at
new coordinates $\hat{x}_1,\dots,\hat{x}_\ell=(\hat{t}_1,1),\dots,(\hat{t}_\ell,1)$. With the notation for a 'generalized' time-series it is easy to insert it to the Transformer/Taylorformer machinery. Note that the additional features for the Taylorformer are calculated separately for the observations and for the climate model outputs since, for instance, we do not want to mix derivatives' values between the two time-series. For full details of the training and inference procedures for BC refer back to the main text. 

\section{Discussion: other ways to slice and infer the data} \label{discussion_slice_infer}
In Algorithm \ref{train_batch_generation} (and the additional specification for multiple initial conditions), we have detailed a specific slicing procedure for the Taylorformer consisting of the steps "Initial condition selection"$\rightarrow$ "Time window selection"$\rightarrow$ "Prediction index selection". But this is just one possibility. We now explain what is the reasoning behind a valid slicing operation.    

\subsection{Slicing discussion}
It is not trivial to slice the pair of the full trajectories of (climate model, observations) sequences. In Figure \ref{training_construction}, the first column shows the full sequences, and the second column shows example slices. The purpose of the slicing operation is to encourage the correction task instead of a prediction task: if we do not slice the data and set each $O^{\star, i} = [O_{t_1}, \dots, O_{t_n}]$ and \mbox{$\textbf{g}^{i, z} = \left[g^{z}_1, \dots g^{z}_n\right]$} containing all climate model outputs just for different initial-condition indexed by $z$, then the ML model will learn to ignore the climate model outputs completely; the observation value would always be the same for all pairs while the climate model output would keep on changing based on which initial-condition $z$ was chosen and hence uninformative. 

Given the understanding that we have to make our pairing (by some slicing) process in an educated manner, how should we pair the climate model(s) with observations? This is only possible if we make some assumptions about the generating process.

Here are a few possibilities for sensible corpus constructions: 
\begin{enumerate}
    \item Construct a dataset comprised of many locations and multiple initial conditions. Each row describes a unique location and is paired with the initial condition index $z$ to give the sequence of (climate model $z$, observations) for all time points.
    \item Construct a dataset of one unique location, multiple initial conditions, and multiple time-windows. Each row describes the same location and is paired with initial condition index $z$ over changing unique time-windows to give the sequence of (climate model $z$, observations).  
\end{enumerate}
\noindent What makes a corpus construction sensible for our task is its ability to do corrections, i.e., use the information provided by the climate models as opposed to ignoring them and just using information based on observations and time points. Intuitively, for the first option, the location is changing from sequence to sequence, and thus a fixed 'rule' that fits to location $A$ using solely the observations will fail when applied on a different location $B$ (assuming $A$ and $B$ are not very similar). For the second option, we have the same location for all sequences. But, a 'fixed' rule will again fail since it has to be appropriate for different time-windows. For example, if one sequence is from 1948 and another is from 1980, the same 'rule' would not work unless the process is stationary (loosely, not changing its properties over time). 

These configurations are selected examples of slicing the data. The question remains: which possibility should be chosen to construct the corpus? A standard ML procedure is to choose the model that performed best on hold-out likelihood and this is how we chose our slicing above. 

\subsection{Inference discussion}
The Taylorformer architecture permits varying-length sequences. This flexibility is useful when training the model since we do not know what the optimal sequence length is when slicing the original sequence into sub-sequences. However, during inference, this flexibility poses a challenge—what should we feed into the trained model to generate the 20 years of values we are interested in?      
We could have used different setups to execute the sampling (generation) process, and not all generation processes are equal. It is a matter of empirical work to choose the generation process. Our choice above is a convenient choice to align loosely with the training procedure. Alternatively, we could inject the entire time-series of a climate model run and all observations from 1948-1988 and sample day-by-day values up until 2008. Or, we can use the same inputs to generate in non-chronological order, generate the 1st of January 1989, and then fill in the values in-between before continuing to 1990 and further.  

\section{Additional experimental results}\label{appendix_experiments}
\begin{table}[ht]
\centering
\begin{threeparttable}
\caption{Average MSE and Log-Likelihood for Abuja and Tokyo (1989-2008)}
\begin{tabular}{lcccc}
\hline
Model & MSE (Abuja) & Log-Likelihood (Abuja) & MSE (Tokyo) & Log-Likelihood (Tokyo) \\
\hline
IPSL \cite{Lurton-ipsl} & 11.12 & -2.62 & 39.57 & -3.26 \\
Mean-shift \cite{Xu} & 5.86 & -2.30 & 14.45 & -2.75 \\
mean + variance \cite{Ho2012} & 7.23 & -2.41 & 12.29 & -2.67\\
EQM \cite{panofsky1968some} & 7.69 & -2.44 & 12.23 & -2.67 \\
dTsmbc \cite{Robin2021IsTA} & 21.55 & -2.95 & 13.28 & -2.71 \\
EC-BC \cite{Vrac2015MultivariateIntervariableSA} & 24.01 & -3.01 & 129.41 & -3.85 \\
\rowcolor{gray!30} \textbf{Temporal BC} & 3.57 & -1.78 & 6.63 & -2.41 \\
\hline
\end{tabular}
\begin{tablenotes}
\small
\item \textbf{Model Performance Comparison (1989-2008):} This table presents the average Mean Squared Error (MSE) and log-likelihood for various models, based on 32 initial condition runs, assessing performance in Abuja and Tokyo. Lower MSE values indicate better performance, while higher log-likelihood values are preferable. Our Temporal Bias Correction (BC) Machine Learning model is highlighted in gray. For models other than our Temporal BC, the probability distribution is assumed to be Normal, with mean $\mu_{i, model}$ (the model's output at time $t_i$) and a constant variance $\sigma_{model}^{2}$. Here, 
$\sigma_{model}^{2} = \dfrac{1}{N}\sum_{i=1}^{n} \left(\mu_{{i}, model} - o_{i}  \right)^{2}$).  
\end{tablenotes}
\end{threeparttable}
\label{tab:abuja_tokyo_results}
\end{table}

\begin{figure}[H]
    \centering
    \includegraphics[width=\linewidth]{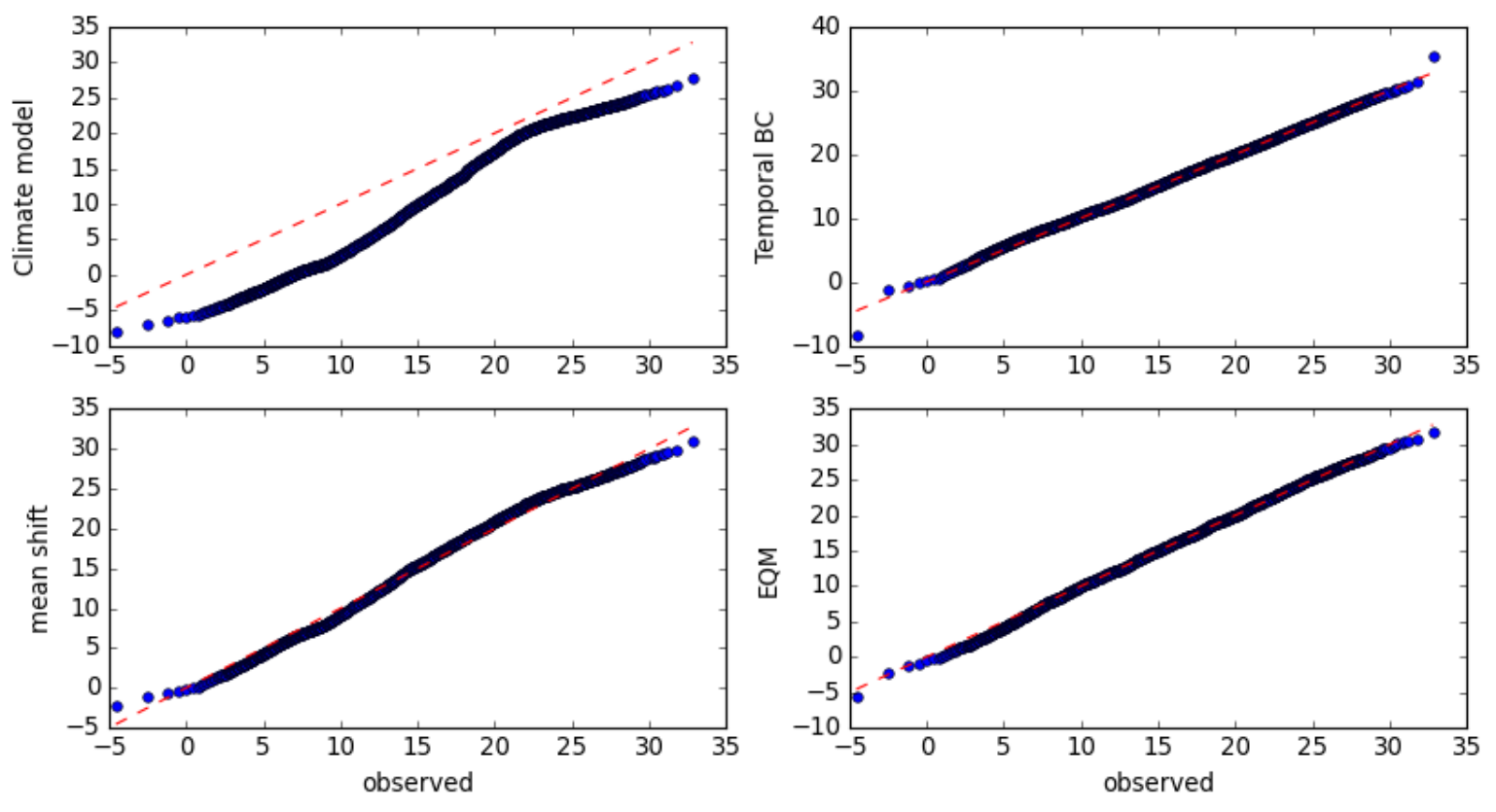}
    \caption[QQ plots for Tokyo, Japan 1989-2008 with different BC methods]{QQ plots for Tokyo, Japan 1989-2008 for maximum daily temperatures (Tmax) for observations (X-axis) versus BC methods (Y-axis): The climate model (IPSL) quantiles (top left plot) are very different than observations (ERA5). For mean-shift (bottom left), EQM (bottom right) and our method \mbox{\textbf{Temporal BC}} (top right) the qauntiles are captured quite well with minor differences}
    \label{QQ_methods_tokyo}
\end{figure}

\begin{figure}[H]
    \centering
    \includegraphics[width=\linewidth]{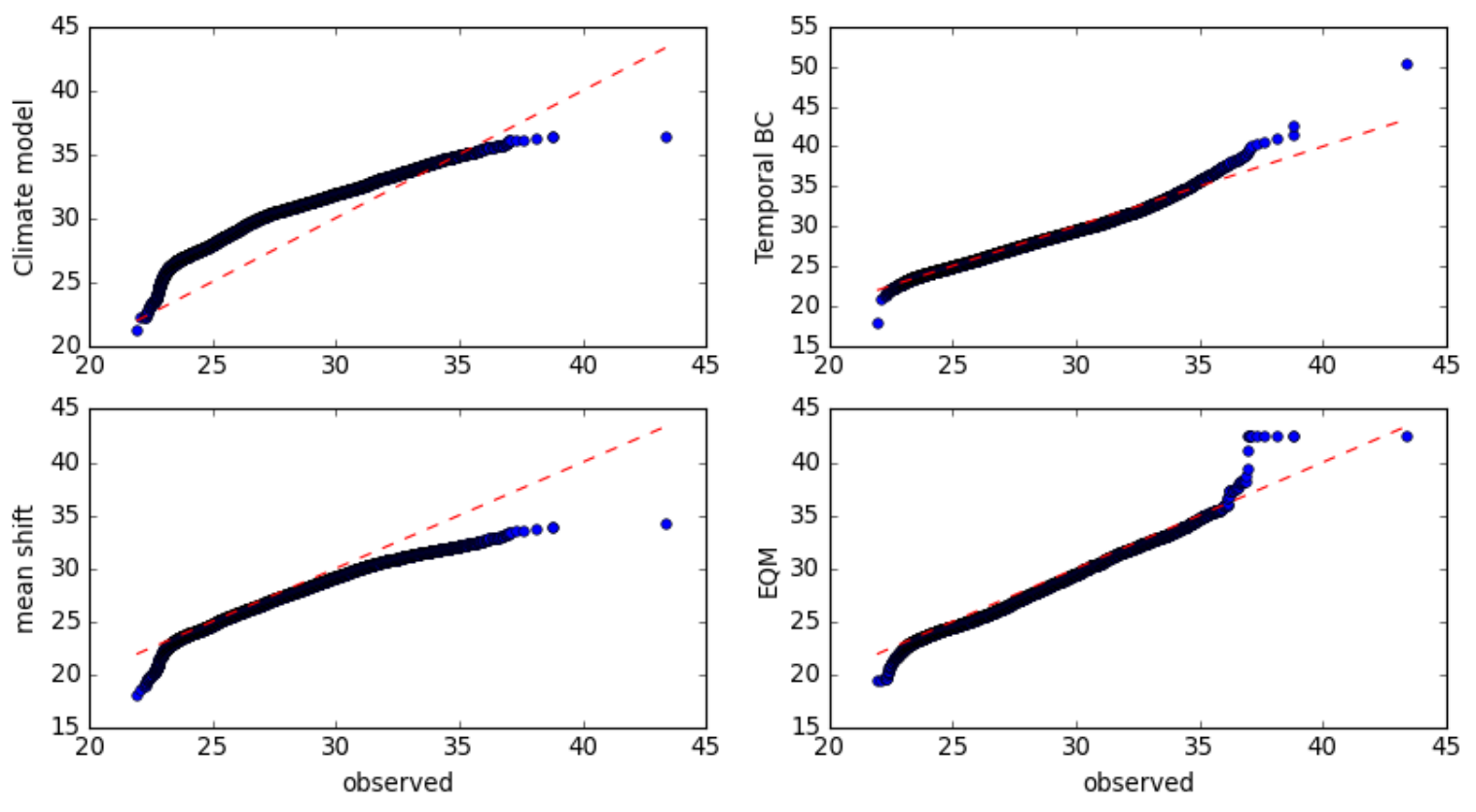}
    \caption[QQ plots for Abuja, Nigeria 1989-2008 with different BC methods]{QQ plots for Abuja, Nigeria 1989-2008 for maximum daily temperatures (Tmax) for observations (X-axis) versus BC methods (Y-axis): The climate model (IPSL) quantiles (top left plot) are very different than observations. For mean-shift (bottom left) and EQM (bottom right) the quantiles are captured more accurately but our method \mbox{\textbf{Temporal BC}} (top right) seems to be the most accurate}
    \label{QQ_methods_abuja}
\end{figure}

\begin{figure}[H]
    \centering
    \begin{minipage}{0.48\textwidth}
        \centering
        \includegraphics[width=\linewidth]{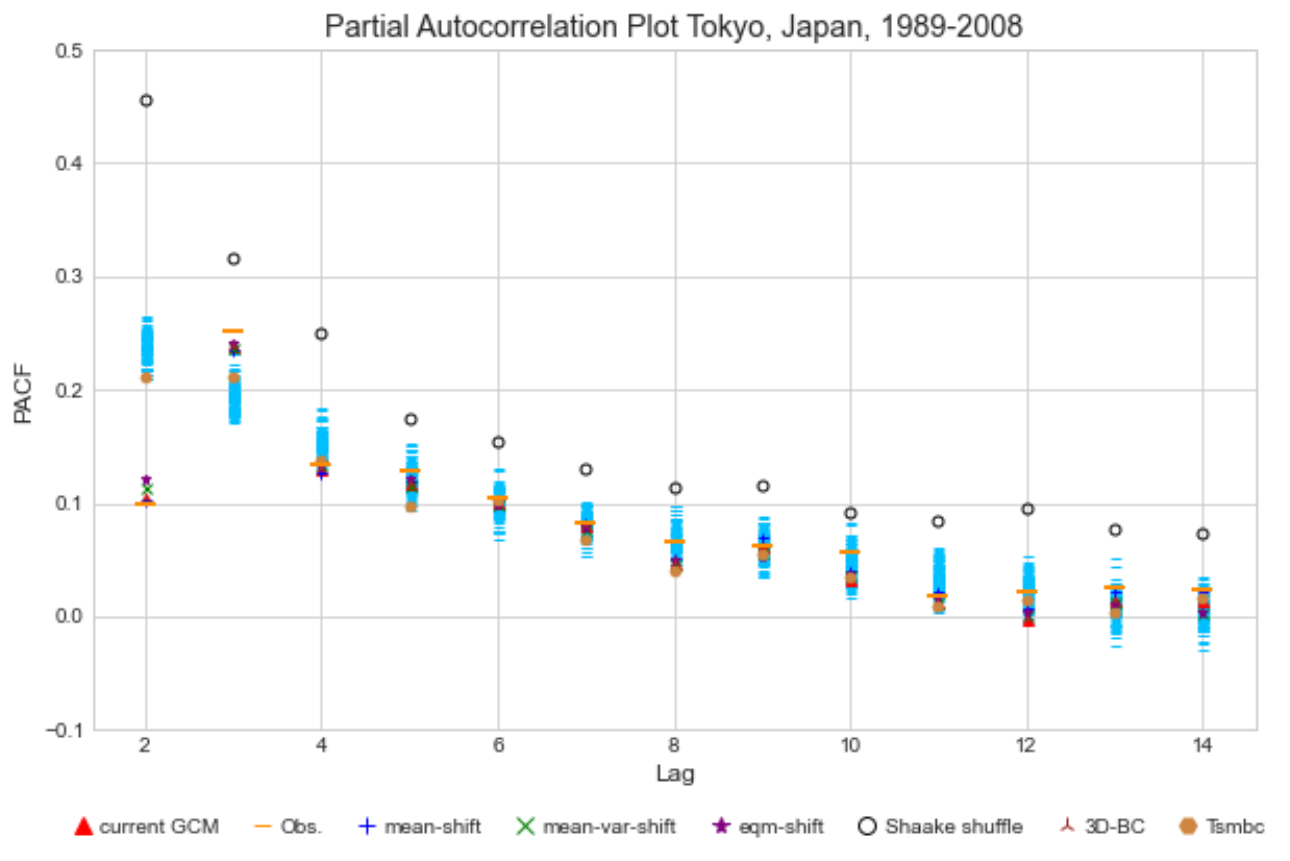}
        \caption[]{Partial auto-correlation plot for Tokyo, Japan, 1989-2008 for lags 2-14 and for a randomly chosen initial condition run. Our Taylorformer temporal BC (horizontal blue lines) outputs multiple trajectories and, therefore, multiple partial auto-correlation values at each lag; it seems to underestimate the observed partial auto-correlation (horizontal orange line) for lags 2 and 3 and capture it well for higher lags. See the markers at the bottom of the figure to interpret the partial auto-correlation for other BC methods.       
        }
        \label{04_05_24_pacf_tokyo_run_11}
    \end{minipage}\hfill
    \begin{minipage}{0.48\textwidth}
        \centering
        \includegraphics[width=\linewidth]{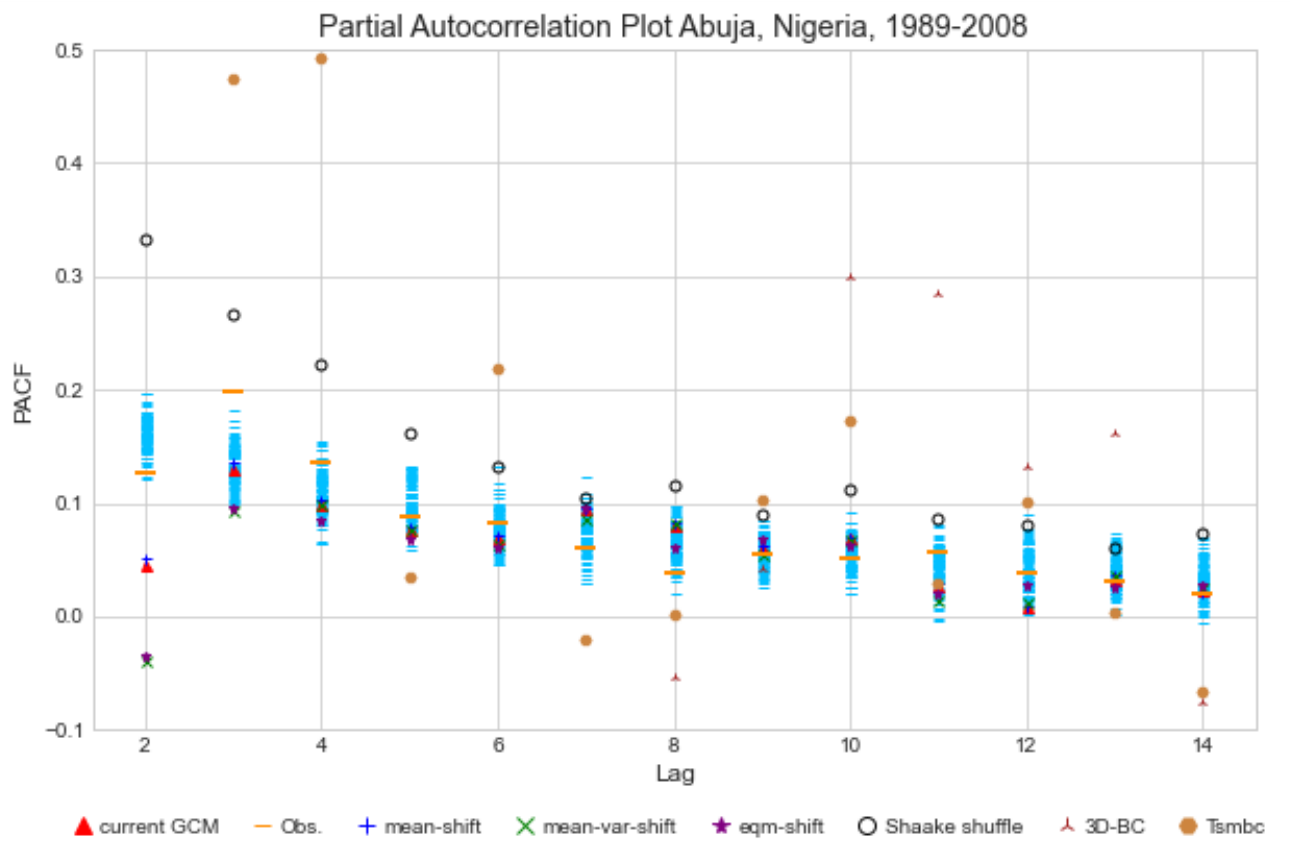}
        \caption[]{Partial auto-correlation plot for Abuja, Nigeria, 1989-2008 for lags 2-14 and for a randomly chosen initial condition run. Our Taylorformer temporal BC (horizontal blue lines) outputs multiple trajectories and, therefore, multiple partial auto-correlation values at each lag; it seems to capture the observed partial auto-correlation (horizontal orange line) for all lags except lag 3. See the markers at the bottom of the figure to interpret the partial auto-correlation for other BC methods.}
        \label{4_05_24_PACF_abuja}
    \end{minipage}
\end{figure}

\begin{figure}[H]
    \centering
    \includegraphics[width=\linewidth]{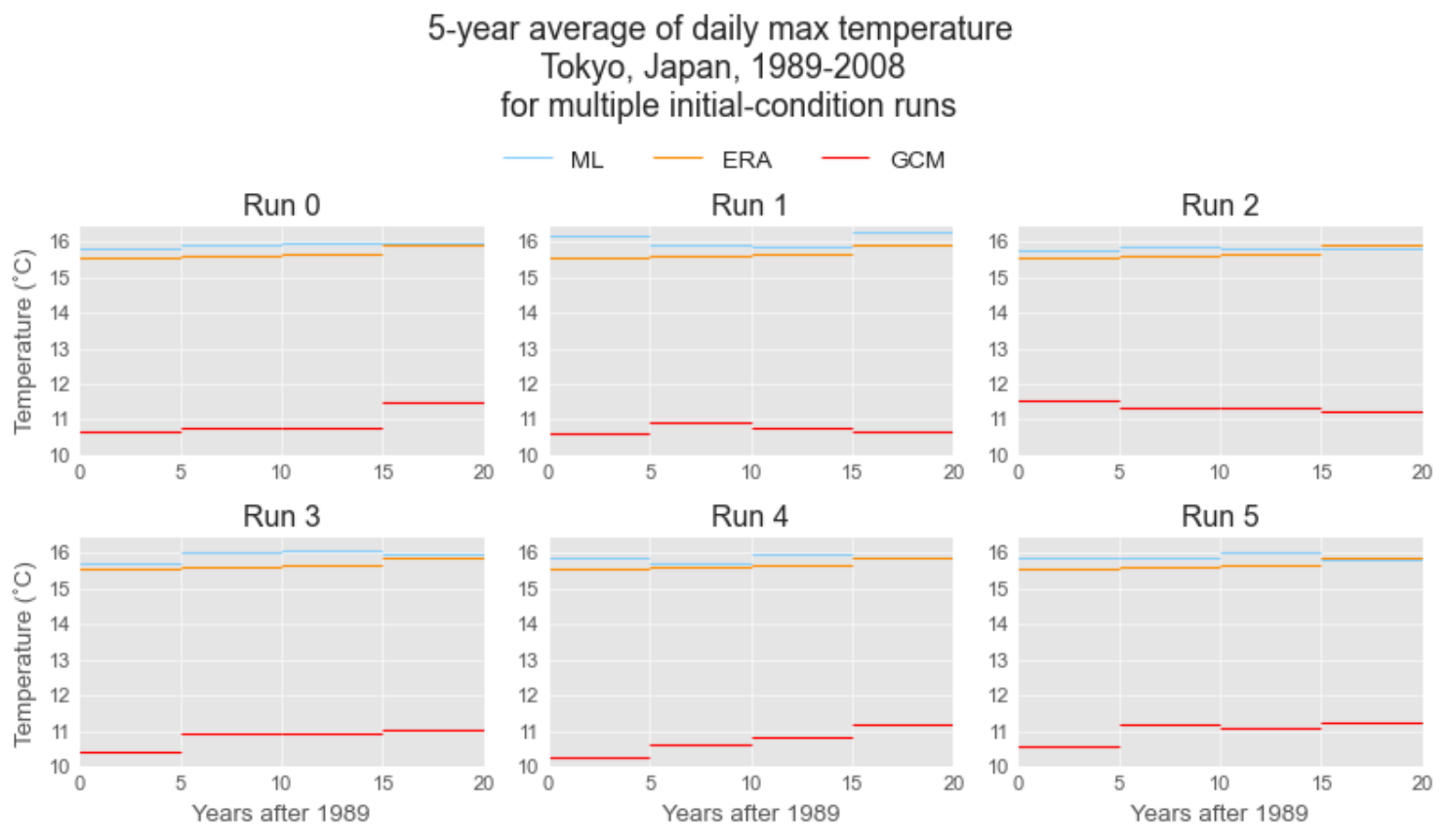}
    \caption[]{\textbf{Comparison of trend evolution of Climate Model, Observations and Temporal BC in Tokyo, Japan (1989-2008):} This figure illustrates the 5-year averages of daily maximum temperatures, with each plot representing a different initial-condition run. The Temporal BC model is depicted by a blue horizontal line, which does not necessarily align with the trend of the initial-condition climate model, shown as a red horizontal line. Furthermore, neither model consistently follows the trend observed in the actual data, represented by an orange horizontal line}
    \label{tokyo_5_year_avg_trend_fig}
\end{figure}

\begin{figure}[ht]
    \centering
    \includegraphics[width=\linewidth]{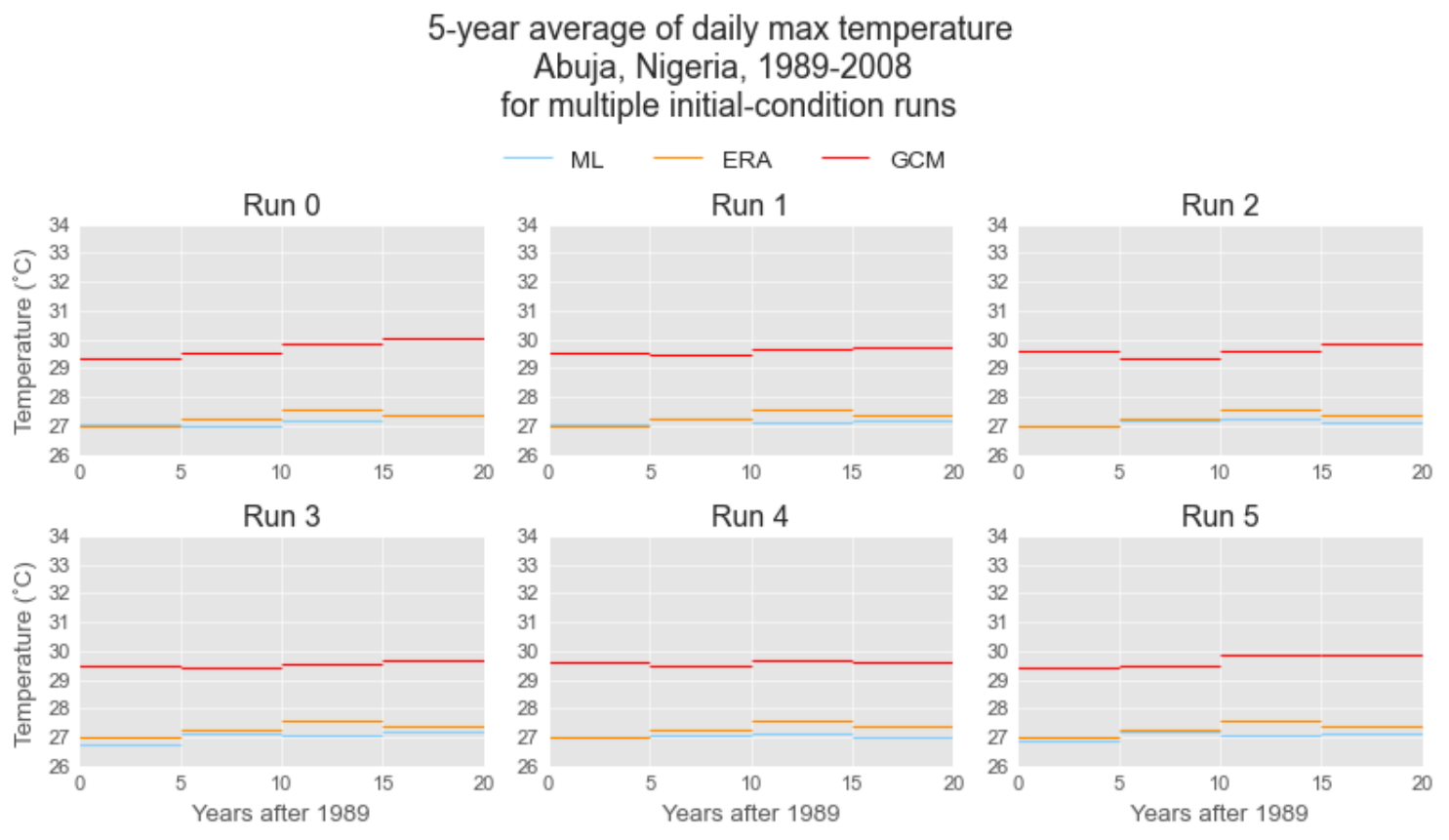}
    \caption[]{\textbf{Comparison of trend evolution of Climate Model, Observations and Temporal BC in Abuja, Nigeria (1989-2008):} This figure illustrates the 5-year averages of daily maximum temperatures, with each plot representing a different initial-condition run. The Temporal BC model is depicted by a blue horizontal line, which does not necessarily align with the trend of the initial-condition climate model, shown as a red horizontal line. Furthermore, neither model consistently follows the trend observed in the actual data, represented by an orange horizontal line}
\label{abuja_nigeria_5_year_avg_trend_preserv_fig}
\end{figure}

\subsection{Tokyo, Japan}
\begin{figure}[H]%
\FIG{\includegraphics[width=0.6\textwidth]{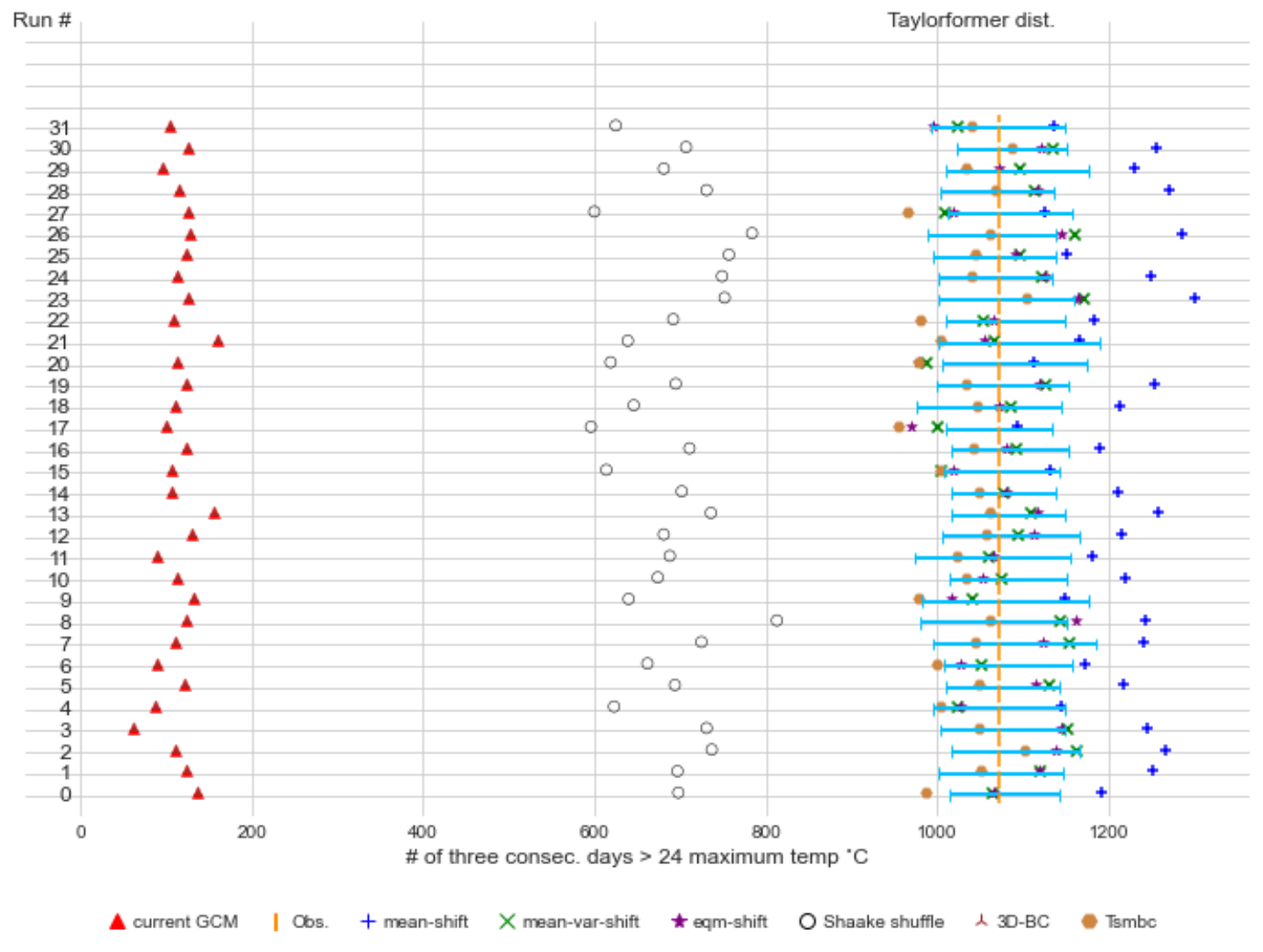}}
{\caption[The raw outputs from the IPSL climate model underestimate...]{\textbf{Comparative Analysis of 'heatwave duration' Trends in Tokyo, Japan (1989-2008):} The number of periods featuring at least three consecutive days with temperatures exceeding 24$^\circ$C is shown. The IPSL climate model predictions are represented by red triangles, which generally underestimate the observations. Actual observations are indicated by a vertical orange line. The Taylorformer temporal BC is depicted using horizontal box plots, with whiskers indicating the 1st and 3rd quartiles. Markers for other BC methods are indicated at the bottom of the figure}   
\label{24}}
\end{figure}

Figure \ref{24} shows the results for Tokyo, Japan using a heatwave threshold of three consecutive days above 24$^\circ$C maximum temperature. The raw outputs from the IPSL climate model (red triangles) underestimate the number of instances of observed heatwaves (vertical orange line) for the period $1989-2008$. Our novel Taylorformer temporal BC produces a more accurate distribution (horizontal box-plots) per run (0-31) for the number of heatwaves in the same period differing on average by $0.7\%$ from the observed number of heatwaves. Most other BC models perform worse: $12.3\%, 4.2\%, 4.1\%, 35.6\%, 89\%, 4.2\%$ for mean-shift, mean and variance shift, EQM, EC-BC, 3D-BC and Tsmbc respectively.

\begin{figure}[H]%
\FIG{\includegraphics[width=0.6\textwidth]{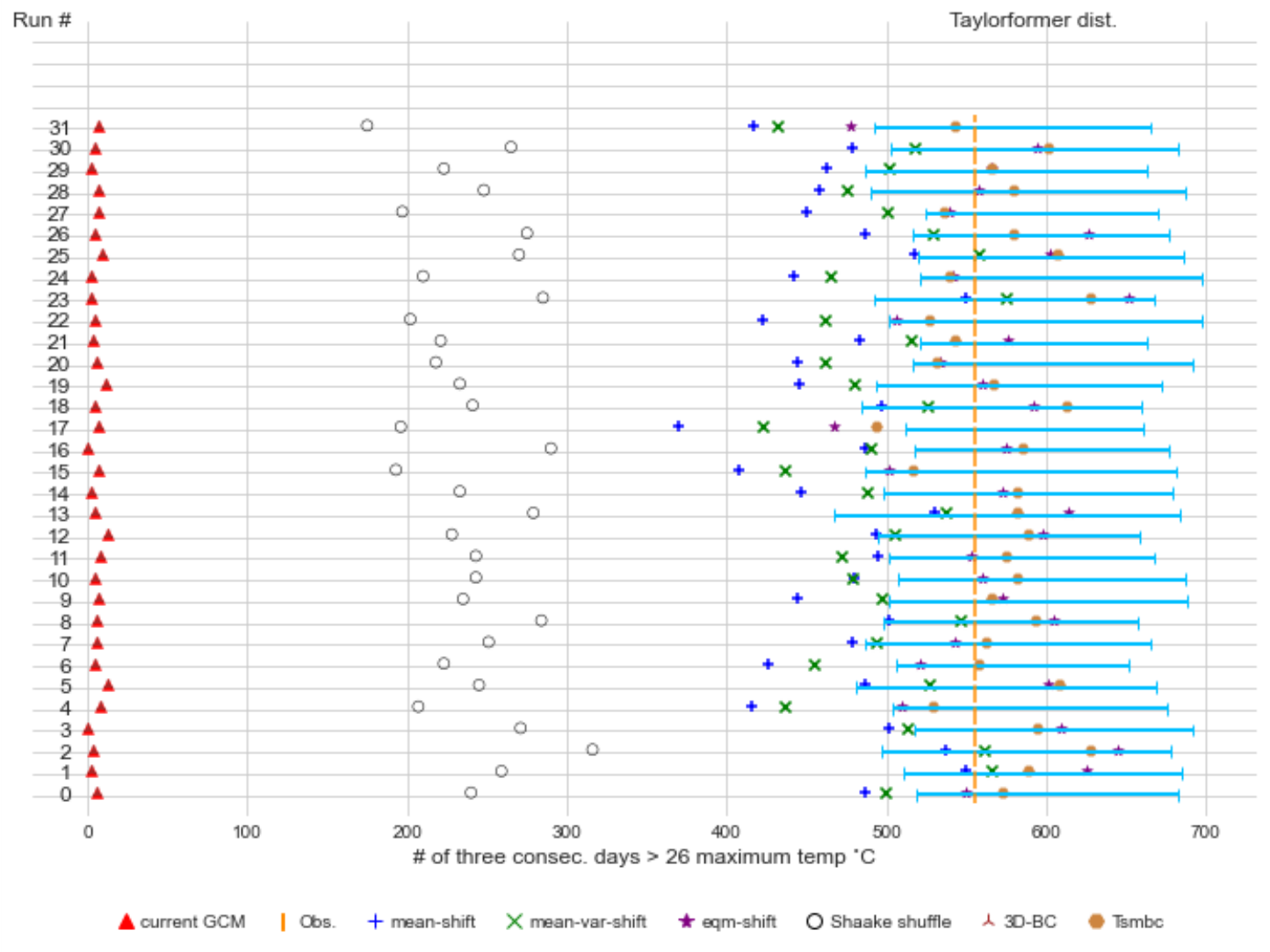}}
{\caption[The raw outputs from the IPSL climate model underestimate...]{\textbf{Comparative Analysis of 'heatwave duration' Trends in Tokyo, Japan (1989-2008):} The number of periods featuring at least three consecutive days with temperatures exceeding 26$^\circ$C is shown. The IPSL climate model predictions are represented by red triangles, which generally underestimate the observations. A vertical orange line indicates actual observations. The Taylorformer temporal BC is depicted using horizontal box plots, with whiskers indicating the 1st and 3rd quartiles. Markers for other BC methods are indicated at the bottom of the figure}     
\label{26}}
\end{figure}

Figure \ref{26} shows the results for Tokyo, Japan using a heatwave threshold of three consecutive days above 26$^\circ$C maximum temperature. The raw outputs from the IPSL climate model (red triangles) underestimate the number of instances of observed heatwaves (vertical orange line) for the period $1989-2008$. Our novel Taylorformer temporal BC produces a more accurate distribution (horizontal box-plots) per run (0-31) for the number of heatwaves in the same period differing on average by $5.6\%$ from the observed number of heatwaves. Most other BC models perform worse: $15\%, 11\%, 7\%, 56\%, 99\%, 5.5\%$ for mean-shift, mean and variance shift, EQM, EC-BC, 3D-BC and TSMBC respectively.

\begin{figure}[H]%
\FIG{\includegraphics[width=0.6\textwidth]{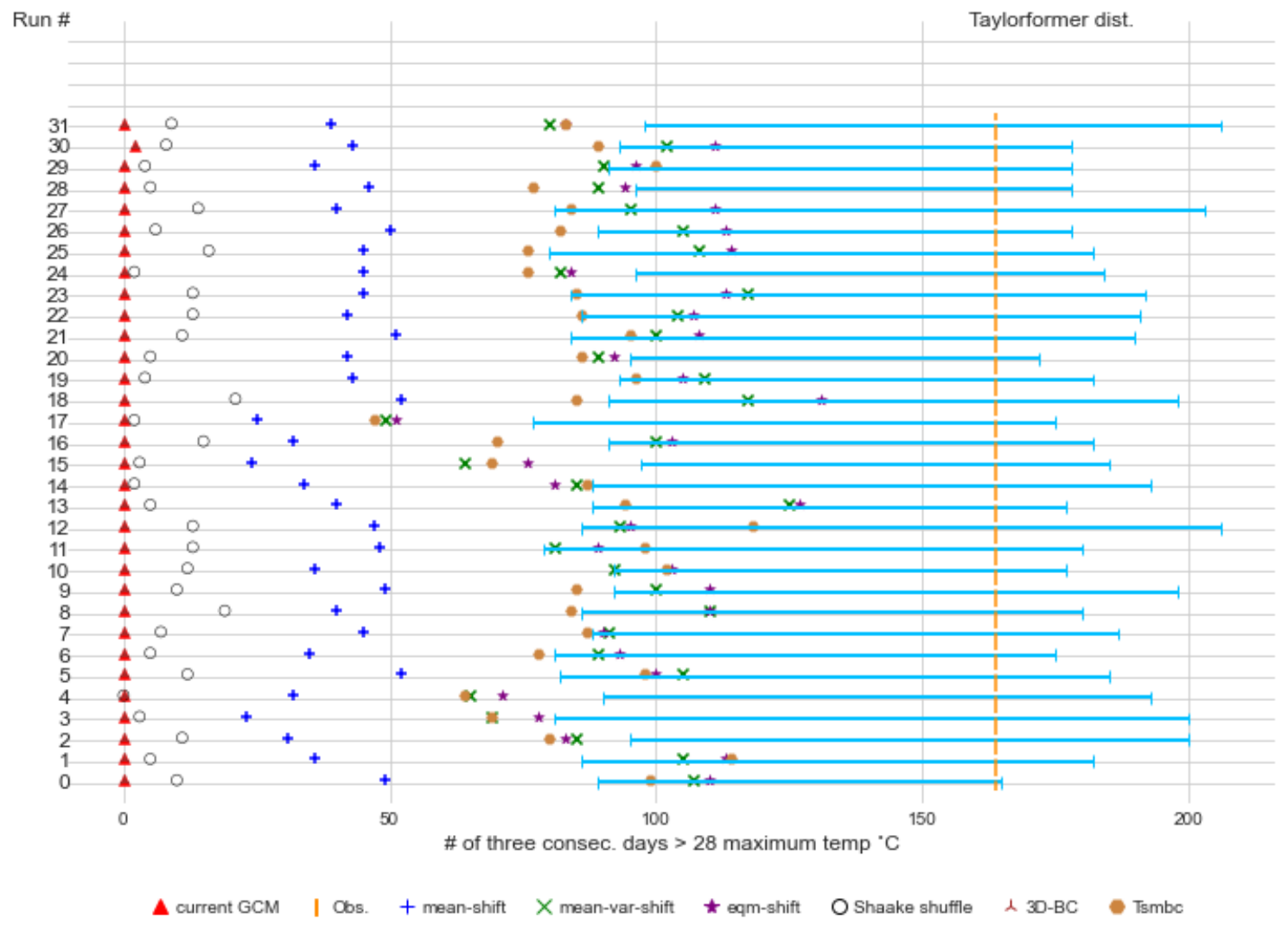}}
{\caption[The raw outputs from the IPSL climate model underestimate...]{\textbf{Comparative Analysis of 'heatwave duration' Trends in Tokyo, Japan (1989-2008):} The number of periods featuring at least three consecutive days with temperatures exceeding 22$^\circ$C is shown. The IPSL climate model predictions are represented by red triangles, which generally underestimate the observations. A vertical orange line indicates actual observations. The Taylorformer temporal BC is depicted using horizontal box plots, with whiskers indicating the 1st and 3rd quartiles. Markers for other BC methods are indicated at the bottom of the figure}  
\label{28}}
\end{figure}

Figure \ref{28} shows the results for Tokyo, Japan using a heatwave threshold of three consecutive days above 28$^\circ$C maximum temperature. The raw outputs from the IPSL climate model (red triangles) underestimate the number of instances of observed heatwaves (vertical orange line) for the period $1989-2008$. Our novel Taylorformer temporal BC produces a much more accurate distribution (horizontal box-plots) per run (0-31) for the number of heatwaves in the same period differing on average by $18.8\%$ from the observed number of heatwaves. Other BC models perform worse: $75.3\%, 42.8\%, 40\%, 94.7\%, 99\%, 47.7\%$ for mean-shift, mean and variance shift, EQM, EC-BC, 3D-BC and Tsmbc respectively.

\subsection{Abuja, Nigeria}
\begin{figure}[H]%
\FIG{\includegraphics[width=0.6\textwidth]{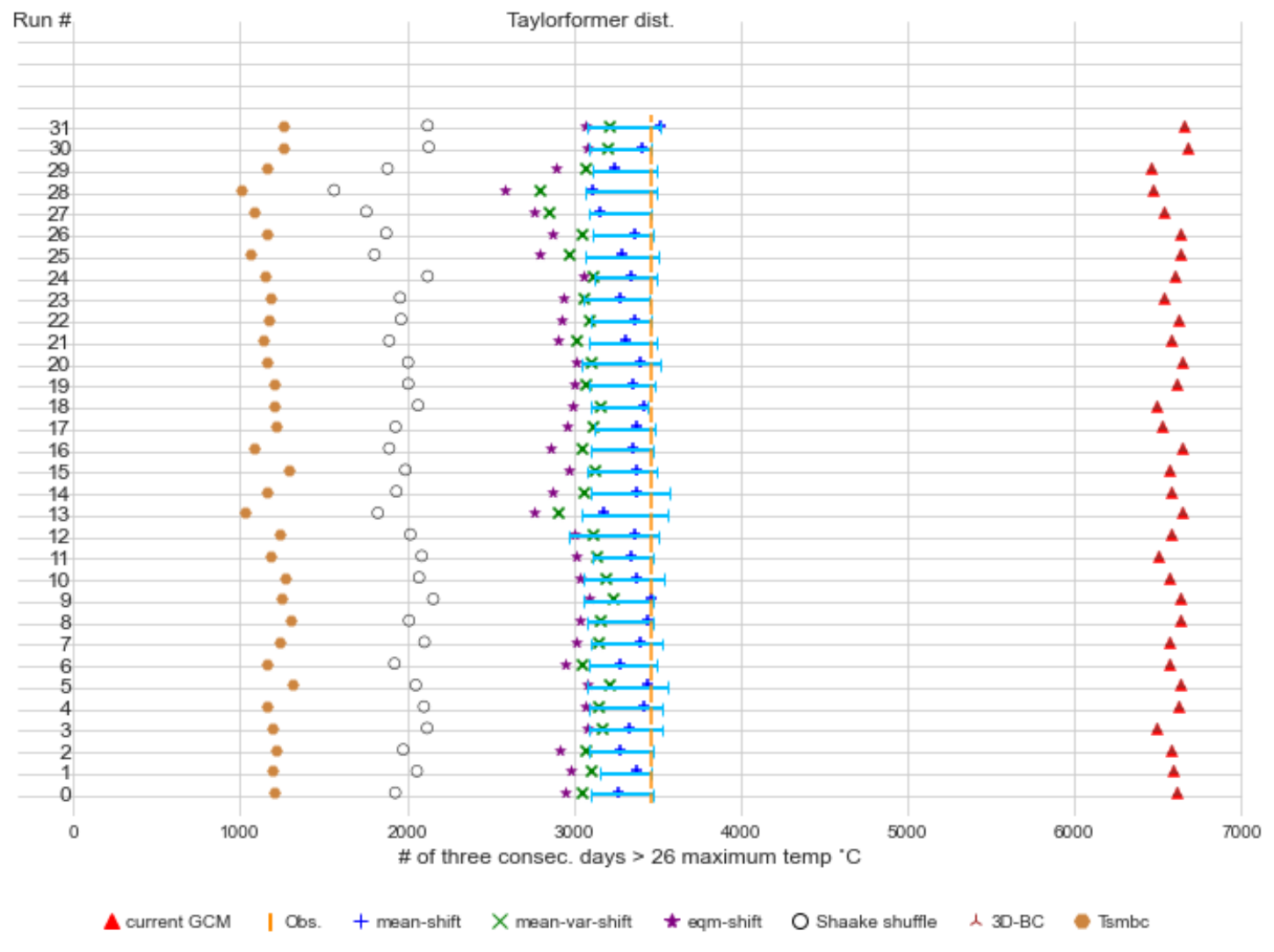}}
{\caption[The raw outputs from the IPSL climate model underestimate...]{\textbf{Comparative Analysis of 'heatwave duration' Trends in Abuja, Nigeria (1989-2008):} The number of periods featuring at least three consecutive days with temperatures exceeding 26$^\circ$C is shown. The IPSL climate model predictions are represented by red triangles, which generally overestimate the observations. A vertical orange line indicates actual observations. The Taylorformer temporal BC is depicted using horizontal box plots, with whiskers indicating the 1st and 3rd quartiles. Markers for other BC methods are indicated at the bottom of the figure}   
\label{abuja_26}}
\end{figure}

Figure \ref{abuja_26} shows the results for Abuja, Nigeria, using a heatwave threshold of three consecutive days above 26$^\circ$C maximum temperature. The raw outputs from the IPSL climate model (red triangles) underestimate the number of instances of observed heatwaves (vertical orange line) for the period $1989-2008$. Our novel Taylorformer temporal BC produces a more accurate distribution (horizontal box-plots) per run (0-31) for the number of heatwaves in the same period differing on average by $4.9\%$ from the observed number of heatwaves. Most other BC models perform worse: $3.7\%, 11\%, 14.8\%, 42.9\%, 89.8\%, 65.9\%$ for mean-shift, mean and variance shift, EQM, EC-BC, 3D-BC and TSMBC respectively.

\begin{figure}[H]%
\FIG{\includegraphics[width=0.6\textwidth]{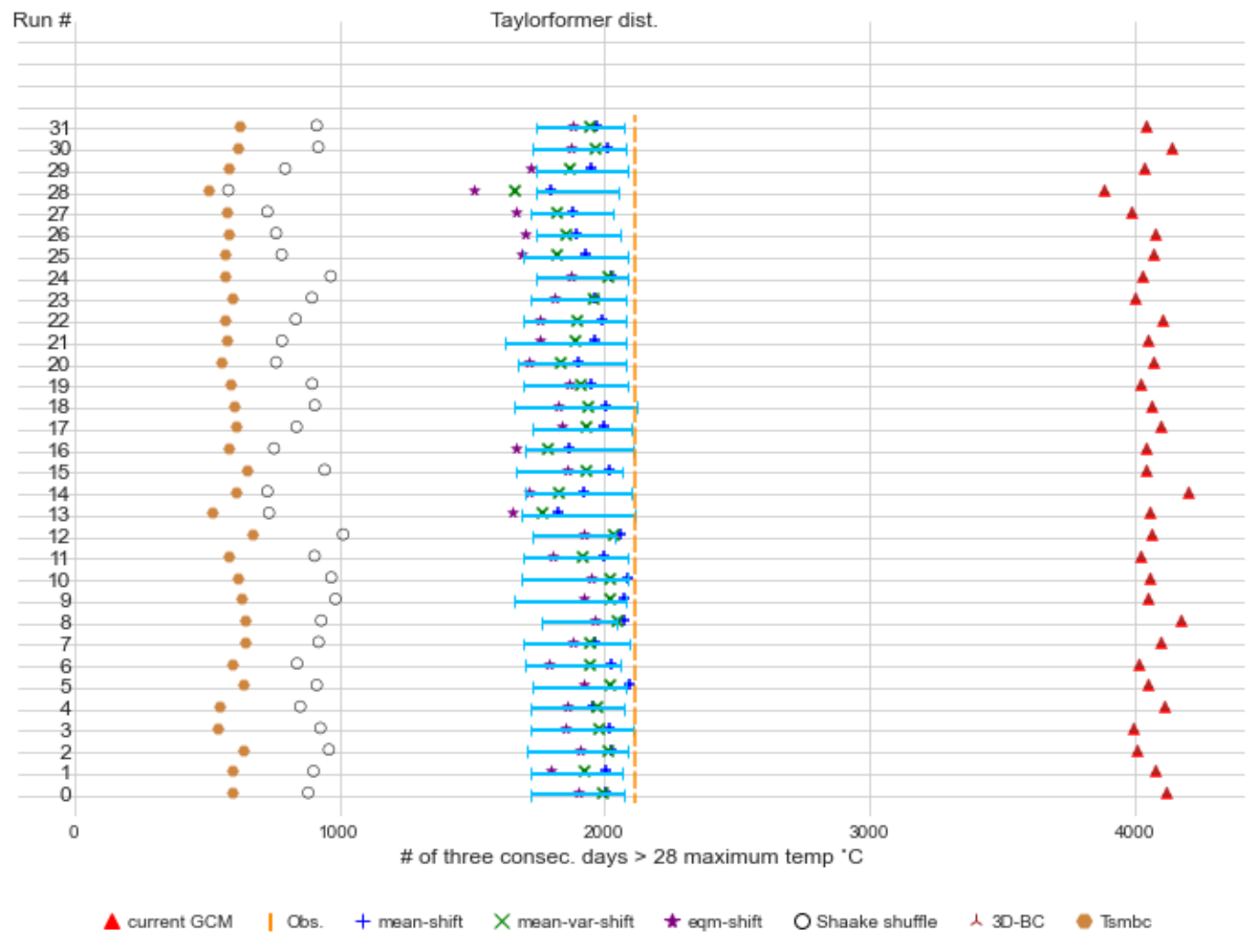}}
{\caption[The raw outputs from the IPSL climate model underestimate...]{\textbf{Comparative Analysis of 'heatwave duration' Trends in Abuja, Nigeria (1989-2008):} The number of periods featuring at least three consecutive days with temperatures exceeding 28$^\circ$C is shown. The IPSL climate model predictions are represented by red triangles, which generally overestimate the observations. A vertical orange line indicates actual observations. The Taylorformer temporal BC is depicted using horizontal box plots, with whiskers indicating the 1st and 3rd quartiles. Markers for other BC methods are indicated at the bottom of the figure}     
\label{abuja_28}}
\end{figure}

Figure \ref{abuja_28} shows the results for Abuja, Nigeria using a heatwave threshold of three consecutive days above 28$^\circ$C maximum temperature. The raw outputs from the IPSL climate model (red triangles) underestimate the number of instances of observed heatwaves (vertical orange line) for the period $1989-2008$. Our novel Taylorformer temporal BC produces a much more accurate distribution (horizontal box-plots) per run (0-31) for the number of heatwaves in the same period differing on average by $10.2\%$ from the observed number of heatwaves. Most other BC models perform worse: $6.5\%, 9.3\%, 14.5\%, 59.3\%, 91.9\%, 71.9\%$ for mean-shift, mean and variance shift, EQM, EC-BC, 3D-BC and TSMBC respectively.

\begin{figure}[H]%
\FIG{\includegraphics[width=0.6\textwidth]{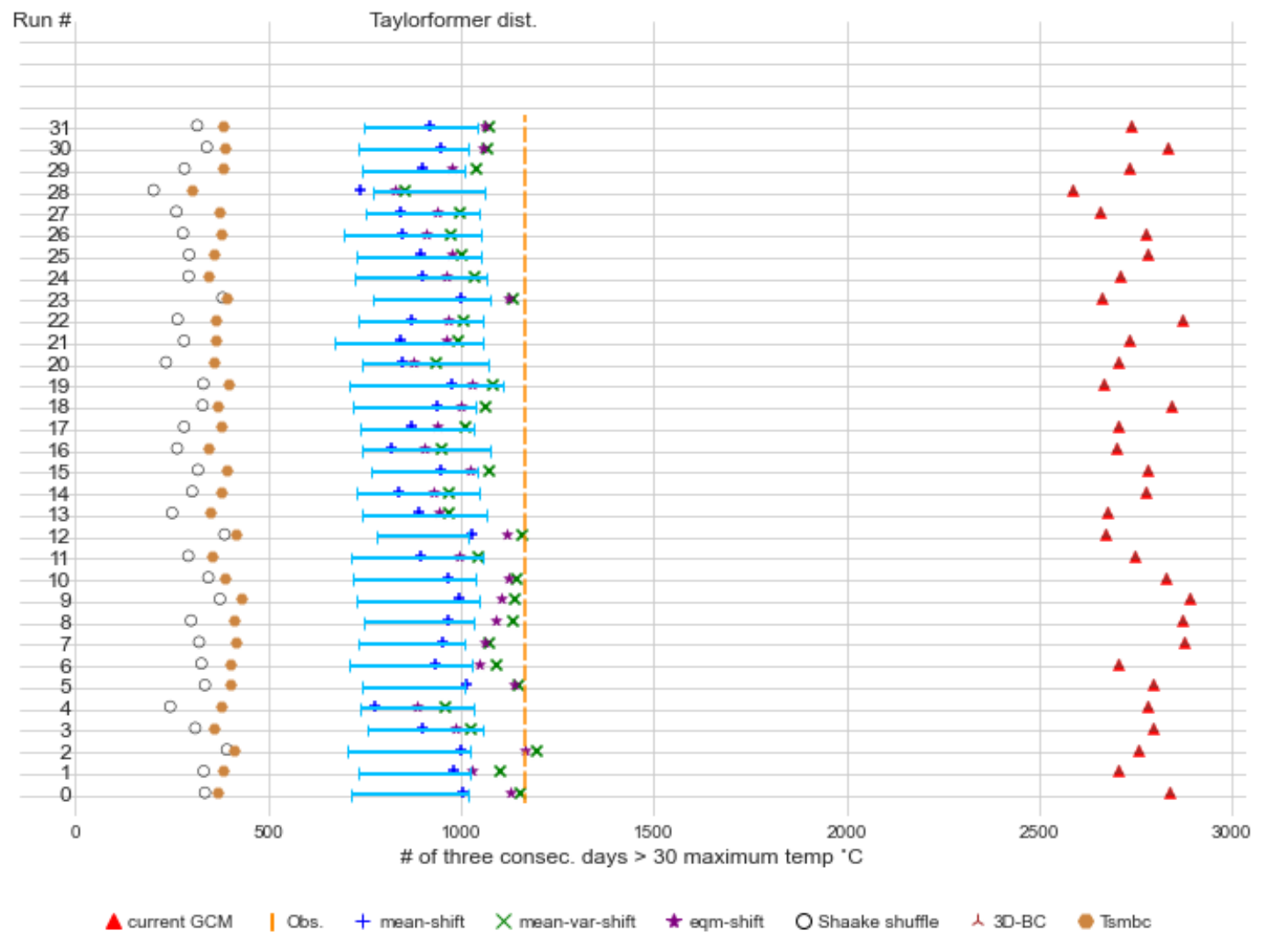}}
{\caption[The raw outputs from the IPSL climate model underestimate...]{\textbf{Comparative Analysis of 'heatwave duration' Trends in Abuja, Nigeria (1989-2008):} The number of periods featuring at least three consecutive days with temperatures exceeding 30$^\circ$C is shown. The IPSL climate model predictions are represented by red triangles, which generally overestimate the observations. A vertical orange line indicates actual observations. The Taylorformer temporal BC is depicted using horizontal box plots, with whiskers indicating the 1st and 3rd quartiles. Markers for other BC methods are indicated at the bottom of the figure}     
\label{abuja_30}}
\end{figure}

Figure \ref{abuja_30} shows the results for Abuja, Nigeria using a heatwave threshold of three consecutive days above 30$^\circ$C maximum temperature. The raw outputs from the IPSL climate model (red triangles) underestimate the number of instances of observed heatwaves (vertical orange line) for the period $1989-2008$. Our novel Taylorformer temporal BC produces a much more accurate distribution (horizontal box-plots) per run (0-31) for the number of heatwaves in the same period differing on average by $23.6\%$ from the observed number of heatwaves. Most other BC models perform worse: $21.7\%, 10.3\%, 13.5\%, 73.7\%, 136\%, 67.6\%$ for mean-shift, mean and variance shift, EQM, EC-BC, 3D-BC and TSMBC respectively.

\section{Supplementary figures}\label{Supplementary_figs}

\begin{figure}[ht]%
\FIG{\includegraphics[width=0.6\textwidth]{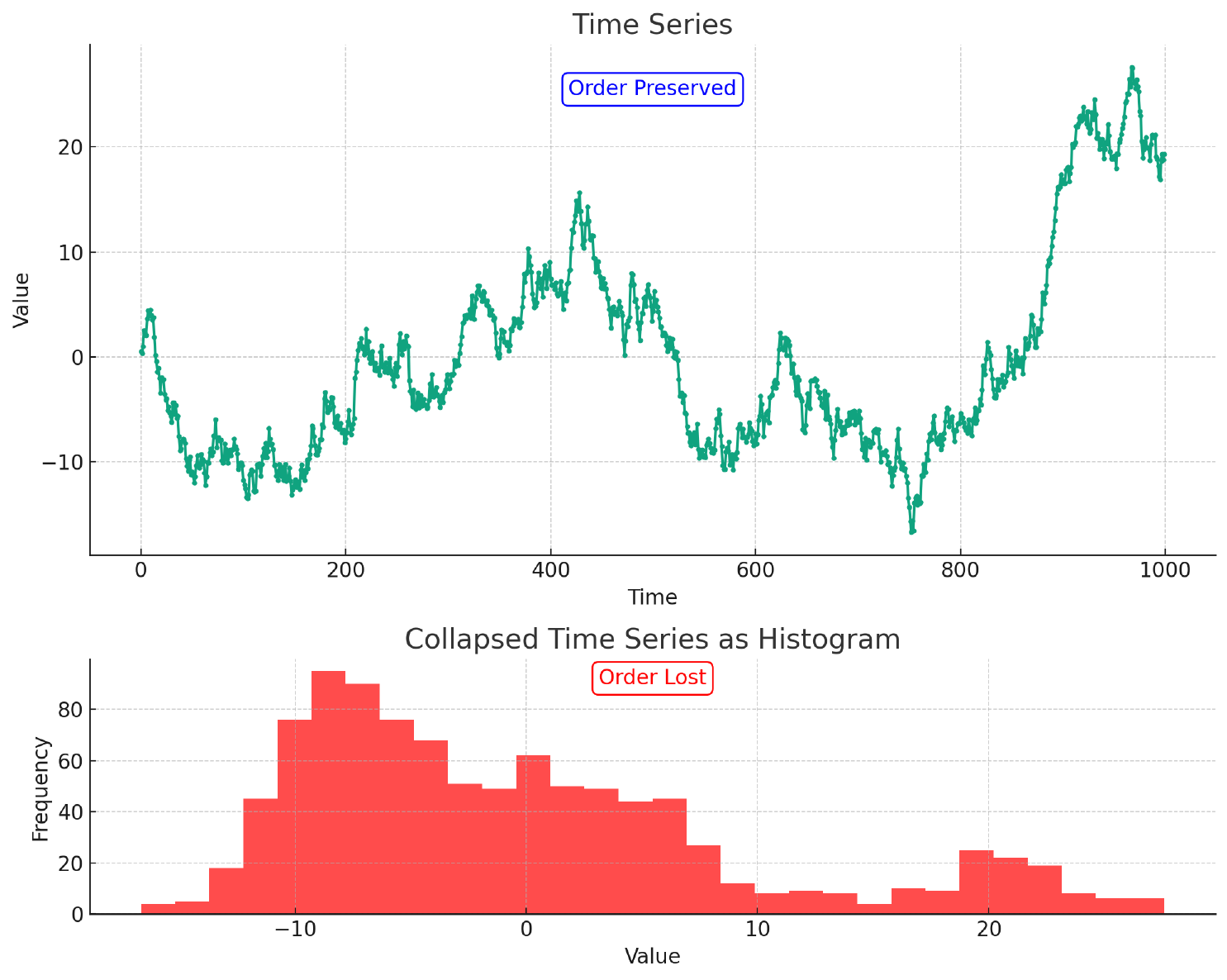}}
{\caption {When collapsing a time-series (top) to a histogram (bottom), the temporal information is lost -- a toy illustration}
\label{gpt_time_collapse}}
\end{figure}

\section{Gaussian Process models}\label{GP_appendix}

\paragraph{Gaussian process models used}
For a time-series with random variables $Y_1, \dots Y_N$ for time-points $t_1, \dots t_N$ the GP in Figure \ref{GP_time_is_meaningful} can be written as: 
\[
Y_i = Z_i  
\]
\noindent Where $Z_i$ is the $i$-th element from the vector $Z \sim \operatorname{MVN}(0, K)$, with $K_{ij} = \exp^{-\dfrac{(t_i - t_j)^{2}}{2\ell^{2}}}$ (this is also known as an RBF kernel) $\ell$ is called the length-scale and for the figure we took $\ell = 0.1$. 
\newline \newline
The GP in Figure \ref{GP_seasonal} can be written as:
\[
Y_i = Z_i * W_i
\]
\noindent Where $Z_i$ is the $i$-th element from the vector $Z \sim \operatorname{MVN}(0, K)$, with $K_{ij} = \exp^{-0.5 * (\sin{(\pi r)/\gamma})^{2}/\ell^{2}}$ (this is also known as a Periodic kernel) $r, \ell, \gamma$ is the euclidean distance between $t_i$ and $t_j$, $\ell$ is the length-scale and $\gamma$ is the period parameter, all set to one. $W_i$ is the $i$-th element from the vector $W \sim \operatorname{MVN}(0, K)$, with $K_{ij} = (\dfrac{1 + r^{2}}{2* \alpha * \ell^{2}})^{-\alpha}$ (this is also known as a Rational Quadratic kernel). Here, $\alpha$ determines the relative weighting of small-scale and large-scale fluctuations. 
\newline \newline
And the GP in Figure \ref{time_misplacement_fig}
can be written as:
\[
Y_{i,k} = Z_i + E_k
\]
\noindent  Where $Z_i$ is the $i$-th element from the vector $Z \sim \operatorname{MVN}(0, K)$, with $K_{ij} = \exp^{-\dfrac{(t_i - t_j)^{2}}{2\ell^{2}}}$ (this is also known as an RBF kernel). $E_k \sim \operatorname{Normal}(0, 2)$, with $k=1$ representing 'observations' and $k=2$ representing 'climate model' outputs. Note that the time mismatch is created by shifting 'observations' timestamps by $0.9$ so that for $k=2$ we have time-points $\mathbf{t} = (t_1, \dots t_N)$ and for $k=1$ we have $\mathbf{t} + 0.9$  

\section{BC models -- implementation details}\label{implementation details}
We have compared our Taylorformer to six BC models. We detail their implementations below:

\paragraph{Mean-shift}
Calculate the difference in means between the climate model and observations for each month and initial condition run over the reference period ($1948-1988$). Then, apply the monthly difference to the climate model in the same month during the projection period ($1989-2008$).

\paragraph{Mean and variance shift}
For each month and climate model initial condition run over the reference period ($1948-1988$), calculate the means and standard deviations (std) of the climate model and separately for observations. Then, for each climate model value in the same month over the projection period ($1989-2008$), subtract the climate model mean of the reference period, multiply by the ratio of observation std to climate model std and finally add the observed mean during the reference period. (Our implementation follows Hawkins et al.\cite{Hawkins2013CalibrationAB}.)

\paragraph{Empirical Quantile Mapping}
For each month and climate model initial condition run over the reference period ($1948-1988$), sort the observations and separately the climate model outputs. For the unsorted values during the same month during the projection period, find the index of the first value bigger or equal in the sorted climate model data (during reference). The corrected value is then the observed value at that index.     

\paragraph{EC-BC}
First, perform EQM as detailed above. Then, for each climate model initial condition run, get the ranks of observations during the reference period ($1968-1988$). Use these ranks to re-sort the climate model outputs during the projection period ($1989-2008$).   

\paragraph{3D-BC}
For each climate model initial condition run we follow the exact algorithm provided in the paper by Mehrotra et al. \cite{Mehrotra2019ARA}. The reference period is $1968-1988$ and the projection period is 1989-2008. 

\paragraph{TSMBC}
Here, we have used the R package 'SBCK' \cite{SBCK}. We initialized it with the code: \texttt{dTSMBC\$new(lag = 10)}. Then, for each climate model initial condition run, we used the code \texttt{tsmbc\$fit(Y0, X0, X1)}
followed by \texttt{tsmbc\$predict(X1)}, where \(Y0\) stands for observations during the reference period, and \(X0, X1\) for climate model outputs in the reference and projection period, respectively.
\end{appendix}

\begin{Backmatter}

\paragraph{Funding Statement}
This research was supported by the Artificial Intelligence for Environmental Risks (AI4ER) CDT, University of Cambridge. Mathieu Vrac's work is also supported by state aid managed by the National Research Agency under France 2030 bearing the references ANR-22-EXTR-0005 (TRACCS-PC4-EXTENDING project) and ANR-22EXTR-0011 (TRACCS-PC10-LOCALISING project).   

\paragraph{Competing Interests}
None

\paragraph{Data Availability Statement}
All data and code is open-sourced and available:
IPSL climate model (data) -- \url{https://esgf-node.llnl.gov/search/cmip6/}, NCEP-NCAR Reanalysis 1 (data) -- \url{https://psl.noaa.gov/data/gridded/data.ncep.reanalysis.html},
ERA5 (data) \url{https://www.ecmwf.int/en/forecasts/dataset/ecmwf-reanalysis-v5}, Taylorformer temporal BC model (code) --\url{https://github.com/oremnirv/Taylorformer}

\paragraph{Ethical Standards}
The research meets all ethical guidelines, including adherence to the legal requirements of the study country.

\paragraph{Author Contributions}
\textbf{Omer Nivron}:  Conceptualization, Methodology, Writing- Original draft preparation, Investigation, Visualization, Software, Formal analysis. \textbf{Damon Wischik}: Conceptualization, Formal analysis, Methodology, Writing- Reviewing and Editing. \textbf{Mathieau Vrac}:  Formal analysis, Writing- Reviewing and Editing, Validation, Supervision. \textbf{Emily Shuckburgh}: Writing- Reviewing and Editing, Supervision. \textbf{Alexander Archibald}: Writing- Reviewing and Editing, Supervision.    

\paragraph{Provenance Statement}
This article is part of the Climate Informatics 2024 proceedings and was accepted in Environmental Data Science on the basis of the Climate Informatics peer review process.
\paragraph{Supplementary Material}
Additional figures and implementation algorithms are provided in appendices A-I.
\printbibliography
\end{Backmatter}
\end{document}